\documentclass[journal]{IEEEtran}
\ifCLASSINFOpdf
\else
\fi
\usepackage{CJK}
\usepackage{indentfirst}
\usepackage{graphicx}
\usepackage{float}
\usepackage{dirtree}
\usepackage{amsmath,amssymb}
\usepackage{setspace}
\usepackage{mathtools}
\usepackage{float}
\usepackage[normalem]{ulem}
\usepackage{multirow}
\usepackage{array}
\usepackage{amsmath}
\usepackage{multicol}
\usepackage{multirow}
\usepackage{cite}
\usepackage{url}
\newcommand{\li}{\uline{\hspace{0.5em}}}
\usepackage{algorithm}
\usepackage{algorithmicx}
\usepackage{algpseudocode}
\usepackage{color}

\begin{document}
\let\sss= \scriptscriptstyle
\let\s= \scriptstyle
\let\ttt= \textstyle
%
\title{User-Guided Personalized Image Aesthetic Assessment based on Deep Reinforcement Learning}

\author{Pei~Lv,
	Jianqi~Fan, Xixi~Nie,
	Weiming~Dong, ~\IEEEmembership{Member,~IEEE},
	Xiaoheng~Jiang,
	Bing~Zhou,
	Mingliang~Xu ~\IEEEmembership{Member,~IEEE},
	and~Changsheng~Xu,~\IEEEmembership{Fellow,~IEEE}
	\thanks{Pei Lv, Jianqi Fan, Xixi Nie, Xiaoheng Jiang, Bing Zhou and Mingliang Xu are with the School of Information Engineering, Zhengzhou University, Zhengzhou 450001, China. E-mail: {ielvpei, iexhjiang, iebzhou, iexumingliang}@zzu.edu.cn; zzu\_fjq@163.com; sunny\_nxx@126.com}
	\thanks{Weiming Dong, and Changsheng Xu are with the NLPR, Institute of Automation, Chinese Academy of Sciences, Beijing 100864, China, and also with the University of Chinese Academy of Sciences, Beijing 100049, China. E-mail: {weiming.dong, changsheng.xu}@ia.ac.cn}

}

\markboth{IEEE TRANSACTIONS ON MULTIMEDIA, ~Vol.~1, No.~1, March~2021}%
{Shell \MakeLowercase{\textit{et al.}}: Bare Demo of IEEEtran.cls for IEEE Journals}
%



\maketitle

\begin{abstract}
Personalized image aesthetic assessment (PIAA) has recently become a hot topic due to its usefulness in a wide variety of applications such as photography, film and television, e-commerce, fashion design and so on. This task is more seriously affected by subjective factors and samples provided by users. In order to acquire precise personalized aesthetic distribution by small amount of samples, we propose a novel user-guided personalized image aesthetic assessment framework. This framework leverages user interactions to retouch and rank images for aesthetic assessment based on deep reinforcement learning (DRL), and generates personalized aesthetic distribution that is more in line with the aesthetic preferences of different users. It mainly consists of two stages. In the first stage, personalized aesthetic ranking is generated by interactive image enhancement and manual ranking, meanwhile two policy networks will be trained. The images will be pushed to the user for manual retouching and simultaneously to the enhancement policy network. The enhancement network utilizes the manual retouching results as the optimization goals of DRL. After that, the ranking process performs the similar operations like the retouching mentioned before. These two networks will be trained iteratively and alternatively to help to complete the final personalized aesthetic assessment automatically. In the second stage, these modified images are labeled with aesthetic attributes by one style-specific classifier, and then the personalized aesthetic distribution is generated based on the multiple aesthetic attributes of these images, which conforms to the aesthetic preference of users better. Once these two stages are completed, our approach can automatically generate personalized aesthetic assessment and ranking of the images provided by the user through comparing with previous mentioned personalized aesthetic distribution. Compared with other existing methods, our approach has achieved new state-of-the-art in the task of personalized image aesthetic assessment on the public AVA and FLICKR-AES datasets. 
\end{abstract}


%
\IEEEpeerreviewmaketitle

\setlength{\parindent}{2em}

\section{Introduction}
%
%
%
%
\IEEEPARstart{P}{ersonalized} image aesthetic assessment is a vital problem that has numbers of applications such as film and television, $etc$. Due to the subjectivity of individual preference and the vast of images, it is challenging for users to immediately outcrop images that satisfy them. In this paper, we aim to model more accurate personalized aesthetic distribution through small amount of user interactions and personalized images, which helps to more quickly find out images that meet the user's aesthetic preferences.

\begin{figure}[t]
\centering
\includegraphics[width=1.0\linewidth]{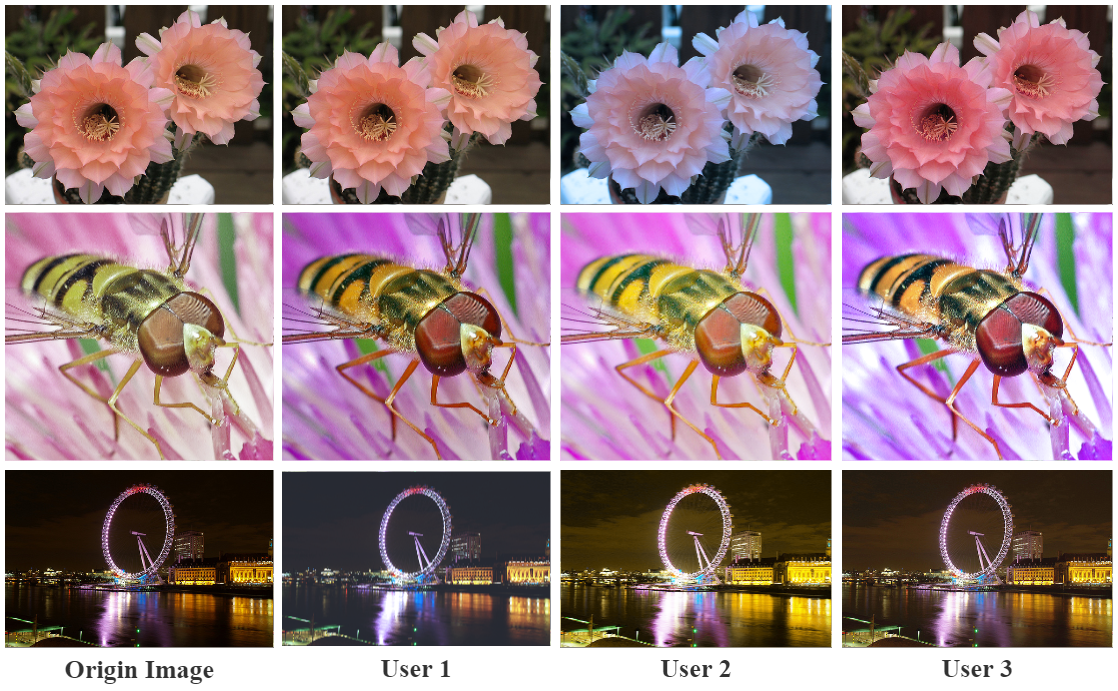}
\caption{Image retouching results by different users. The first column is the original images; the other columns are the results by different users.}
\label{fig:campare_image}
\vspace{-0.55cm}
\end{figure}

In recent years, plenty of work has been devoted to image aesthetic assessment. Some hand-crafted \emph{low-level visual features} \cite{tong2004classification,datta2006studying,luo2008photo,wong2009saliency,li2009aesthetic,nishiyama2011aesthetic} and \emph{high-level aesthetic attributes} \cite{dhar2011high,marchesotti2011assessing,tang2013content} have been proposed to estimate the image aesthetic quality. Although using these traditional aesthetic features to classify images has made some progress, their accuracy can still not satisfy the requirement of different users. It is gratifying that the state-of-the-art features extracted by \emph{deep convolution neural networks} \cite{PIAA2020,lu2014rapid, kong2016photo,lu2015deep,jin2018ilgnet,jin2016image,kang2014convolutional,wang2016brain,mai2016composition,yeh2010personalized,yeh2014personalized,jin2018predicting} have been involved to evaluate the aesthetic quality of images due to its powerful representation. However, a small set of annotated images are insufficient to fully describe the user's personal preference owing to its strong subjectivity. Some methods\cite{PIAA2020,lv2018usar,ren2017personalized,li2020personality} have introduced personalized features into aesthetic assessment. Nevertheless, in these methods, users are only allowed to express limited aesthetic preferences through the existing original images without changing the contents of them. The image retouching results by different users (shown in Fig.~\ref{fig:campare_image}) show that the existing original images can not fully highlight the aesthetic preferences of different users.

\begin{figure}[t]
\centering
\includegraphics[width=1.0\linewidth]{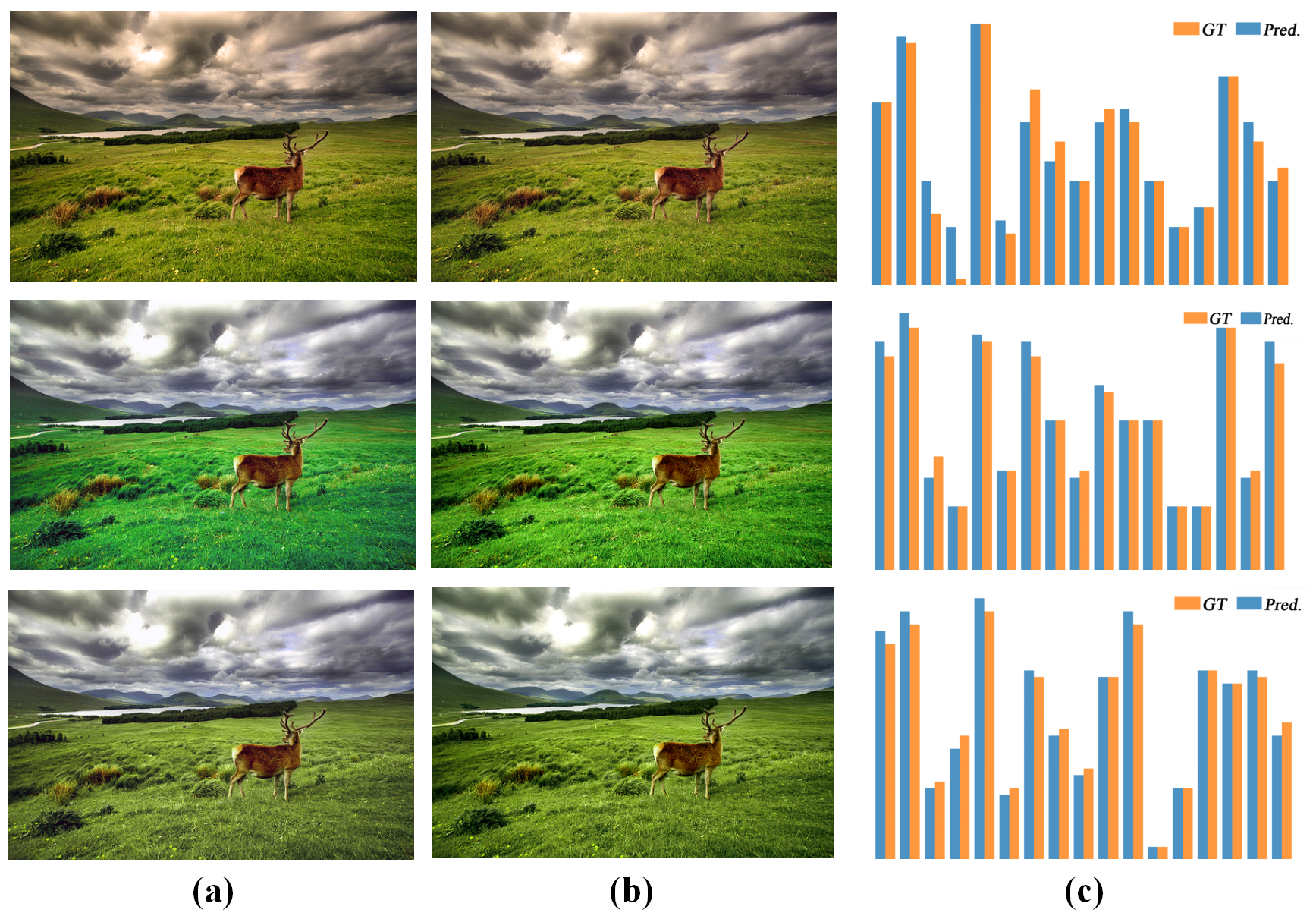}
\caption{The results of personalized image enhancement. (a) ground-truth of retouching results by different users; (b) our automatic retouched results; (c) the personalized aesthetic distributions of different users.}
\vspace{-0.5cm}
\label{fig:fig1}
\end{figure}

Fortunately, deep convolutional neural networks (DCNNs) have shown great promise for image enhancement. Many approaches have been employed for automatic photo enhancement, including \emph{style transfer} \cite{gatys2016image,wang2017multimodal,liao2017visual} and \emph{automatic image enhancement}\cite{gharbi2017deep,park2018distort,yan2014learning,hu2018exposure,ignatov2018wespe,ignatov2017dslr,ying2017new}. The goal of style transfer is that the combined image will be generated via reconstructing the content of one image and the style of another one. Since the publicly available dataset mainly represents general style characteristics, it can not accurately reflect the personalized aesthetic preferences of different users. Automatic image enhancement intends to gain higher quality images based on black-box image enhancer \cite{ignatov2017dslr,ying2017new}, which lacks of the full consideration of the personalized aesthetic attributes. Hence, by retouching the image with specific user interaction, we can obtain more accurate user personalized aesthetics distribution as shown in Fig.~\ref{fig:fig1}. In addition, training one high-quality DCNN requires a great quantity of data, the approach of personalized image enhancement can be used to alleviate the problem of insufficient training samples.

In this paper, we propose one new framework for image aesthetic assessment, which is termed User-Guided Personalized Image Aesthetic Assessment (UG-PIAA). UG-PIAA mainly consists of two stages: user-guided image aesthetic ranking and personalized aesthetic distribution generation. There are two inherent motivations here. On the one hand, most of public image dataset can not highlight the personalized aesthetic preference of different users. At the same time, even if different users retouch one same image, various results will often be obtained as shown in Fig.~\ref{fig:campare_image}. Based on this fact, we introduce personalized image enhancement into the task of image aesthetic assessment, which is not covered by previous work. On the other hand, the task of ranking and retouching are heavily dependent on visual feedback. Common users need to complete these tasks serially and finally output results based on continuous feedback, rather than directly inferring the final result from the input. For aesthetic evaluation, more comprehensive retouching and ranking process needs to take into account different types of operations, as well as the selective usage of these operations. Since above two kinds of operations can naturally be modeled as the corresponding sequential decision-making procedure, deep reinforcement learning is involved to help to complete the image aesthetic assessment.

The pipeline of our UG-PIAA is as following: it firstly uses the public aesthetic dataset to initialize two different policy networks, and then uses the modified images by the user interaction to optimize the personalized policy networks. In the retouching process of user interaction, the original images will be pushed to the image enhancement policy network, while the retouched images by the user will serve as the feedback to this network. After that, the retouched images will be pushed to the image ranking policy network, while the ranked results by the user will be treated as the feedback of the ranking policy network. The above process will be iterated for multiple rounds until the new image sequence generated by the proposed framework meets the user's aesthetic preference. Finally, we use the modified image sequence from stage one to generate the personalized aesthetic distribution. After that, the correlation coefficient between the personalized aesthetics distribution of the enhanced images and that of the original images is obtained. The larger the coefficient, the higher the aesthetic score and ranking of the images. Some examples of the visualization results are shown in Fig.~\ref{fig:fig1}. To evaluate the performance of the proposed approach, we conduct comprehensive ablation studies on two large-scale public datasets through in-depth and insightful experiments. Our major contributions are listed as follow:
\begin{itemize}
	\item We propose a novel user-guided personalized aesthetic assessment framework via a small amount of retouching and ranking interactions.
	\item We introduce the personalized image enhancement incorporating user interaction into the personalized aesthetic ranking, with the aim of highlighting users' personalized aesthetic preferences.
	\item We develop a deep reinforcement learning approach for personalized aesthetic ranking through online fine-tuning by user interactions. Experiments show that our results outperform existing methods in the field of personalized image aesthetic enhancement and ranking.
\end{itemize}

It is to be mentioned that this work extends the preliminary version of our paper\cite{lv2018usar}, which is mainly improved in the following three aspects: \emph{1)} We provide more comprehensive discussion on the introduction of personalized image aesthetics enhancement into the personalized image aesthetics assessment, and make an intensive evaluation of the proposed personalized image aesthetic enhancement. \emph{2)} We use the deep reinforcement learning approach instead of \emph{SVM}$^{rank}$ in the personalized image aesthetics ranking, and quantitatively evaluate its performance. \emph{3)} We perform more ablation studies for the personalized aesthetic assessment which enables an insight understanding of the results. The positive effect of personalized aesthetics enhancement on personalized aesthetic assessment also points towards a promising avenues for future research.

\section{Related Work}

\subsection{Image aesthetics assessment}
Image aesthetics assessment is an important problem in computer vision and computer graphics in recent years. Previous studies mainly focus on \emph{hand-crafted features}\cite{tong2004classification,datta2006studying} and \emph{high-level features}\cite{luo2008photo,wong2009saliency,nishiyama2011aesthetic,dhar2011high,marchesotti2011assessing,ke2006design,bhattacharya2010framework} extracted from high-quality images.  However, hand-crafted features are not comprehensive, it is often unable to fully capture the diversity and aesthetic features of images. Since deep learning can make up for the above defects, significant progress in this research area has been achieved through deep neural networks\cite{PIAA2020,lu2014rapid,lu2015deep,ma2017lamp,talebi2018nima,lv2018usar,kong2016photo,sheng2018attention,mai2016composition,yeh2010personalized,yeh2014personalized,ren2017personalized,DBLP,wang}. Lu \emph{et al.}\cite{lu2015deep} first proposed a double-columned deep convolutional neural network based on AlexNet. Their approach combined global and local features, adopted a two-way network structure, and output the probability of images aesthetic classification. Wang  \emph{et al.}\cite{wang} proposed the first multi-patch method for the prediction of image aesthetic score with the original image aspect ratios being preserved. Kong \emph{et al.}\cite{ma2017lamp} developed a deep network architecture by combining the aesthetic attributes and content information of images to assess aesthetic quality.
\\\indent However, due to the strong subjectivity of image aesthetic assessment, it is difficult to quantify the criteria for image aesthetic. Some studies \cite{PIAA2020,yeh2010personalized,yeh2014personalized,ren2017personalized,lv2018usar} took the subjectivity of aesthetic assessment into consideration. Yeh \emph{et al.}\cite{yeh2010personalized,yeh2014personalized} introduced the concept of personalized ranking and incorporated user feedback into the ranking algorithm to describe the personalized aesthetic style of the user. Vlad \emph{et al.}\cite{DBLP} proposed a structure that could maintain the resolution of the original image and input images of any size for training. Feature maps of different scales are extracted from different layers of the network and merged to train a \emph{CNN} model to predict the image aesthetic score. In order to learn different human cognition of image aesthetics, Zhu \emph{et al.}\cite{PIAA2020} proposed a PIAA method based on Meta-Learning with bilevel gradient optimization. Xu \emph{et al.}\cite{xu2020spatial} exploited the attention mechanism to learn the spatial attention map of image layout  and find the spatial importance in aesthetics. Li \emph{et al.}\cite{li2020personality} introduced personality traits to the task of the image aesthetics assessment and presented a personality-assisted multi-task deep learning framework for both generic and personalized image aesthetics assessment. Lv \emph{et al.}\cite{lv2018usar} proposed to continuously optimize user aesthetic preferences through multiple rounds of simple user interactions. However, during the interaction phase, user can only rerank the images without changing the contents of these images, but simply ranking the images does not highlight personalized aesthetic preferences very well. In order to solve the above problems and acquire better personalized aesthetic distribution, we further involve user retouching and re-ranking operations into the interaction phase.

\subsection{Aesthetic image enhancement}
Image enhancement is dedicated to improving the aesthetics of an image through transforming or altering the image using various methods and techniques. The work in  \cite{bhattacharya2010framework,lv2018usar,gatys2016image,wang2017multimodal,liao2017visual,gharbi2017deep,park2018distort,yan2014learning,hu2018exposure,deng2018aesthetic,fang2017creatism,guo2018automatic,li2018a2,zeng2019reliable,chen2017learning,wei2018good} enhanced images to obtain more satisfying results over the years. For the problem of style transfer, Gatys \emph{et al.}\cite{gatys2016image} applied the DCNN to produce impressive stylization results by decomposing content and style from images. Liao \emph{et al.}\cite{liao2017visual} founded semantically meaningful dense correspondences between two input images and produced reconstructed results of visual style transfer through deep image analogy. Deng \emph{et al.}\cite{deng2018aesthetic} designed one generator for enhancing low-quality images from the perspective of color transformation. Fang \emph{et al.}\cite{fang2017creatism} decomposed the entire task of image enhancement into a series of independent filters, and connected these filters to complete the final goal. From the perspective of image aesthetics, some researchers also improved the quality of images by \emph{automatically cutting}\cite{guo2018automatic,deng2018aesthetic,zeng2019reliable,park2018distort} and \emph{automatic composition}\cite{chen2017learning,wei2018good}. Li \emph{et al.}\cite{li2018a2} proposed a strategy of clipping images to make them more beauty by reinforcement learning.
\\\indent The above researches mainly consider the image enhancement of common aesthetic attributes. Recently, some researchers tried to model the user's personalized preference to produce personalized stylization in the process of automatic image retouching. Unfortunately, existing algorithms do not fully consider personalized issues such as the user's subjectivity. In view of this, we utilize subjective operations in the process of image modification, according to the retouching and reordering of images by users.

\subsection{Reinforcement Learning}
Reinforcement learning is an important branch of machine learning, which is widely used in games, robotics, natural language processing, image enhancement\cite{park2018distort,hu2018exposure}, $etc$. There have been some excellent work in recent years, like deep Q-network (DQN) \cite{mnih2015human}, double Q-learning \cite{van2016deep}, Dueling network\cite{wang2016dueling}.  In reinforcement learning, the core problem\cite{sutton2018reinforcement} is how an agent learns to adapt a possibly complex, initially unknown environment in a sequential trial and error process. It can directly obtain the learning information and update the model parameters by receiving the reward of the action. Park \emph{et al.} \cite{park2018distort} proposed a deep reinforcement learning method for color reinforcement. It regarded color enhancement as Markov Decision Process, and trained an agent to learn the optimal global enhancement sequence in each step.
\\\indent When one professional photographer retouches images in real life, each operation will depend on the visual feedback presented by the current image. The above series of operations can be described as a trial and error process, which is a sequential decision problem. Inspired by this idea, we use reinforcement learning to model sequential decision problems of retouching and ranking. The reinforcement learning algorithm based on the value function doesn't need to change the selection strategy, but it is not suitable for the large-scale reinforcement learning process. The policy-based reinforcement learning algorithm easily converges to local optimal rather than the global one. Therefore, this paper uses the Actor-Critic algorithm, which breaks the limitations of the above two kinds of algorithms. The Actor chooses the next action according to the strategy, and the Critic gives the score according to the action selected by the Actor. We propose a personalized image enhancement module that adapts the retouched images as the input to the Actor-Critic neural network. The Actor network is used to generate enhanced images, and the Critic network is used to score the generated images so that we can obtain a personalized retouched image with the user preferences as a guide.

\begin{figure*}[t]
\centering
\includegraphics[width=0.8\textwidth]{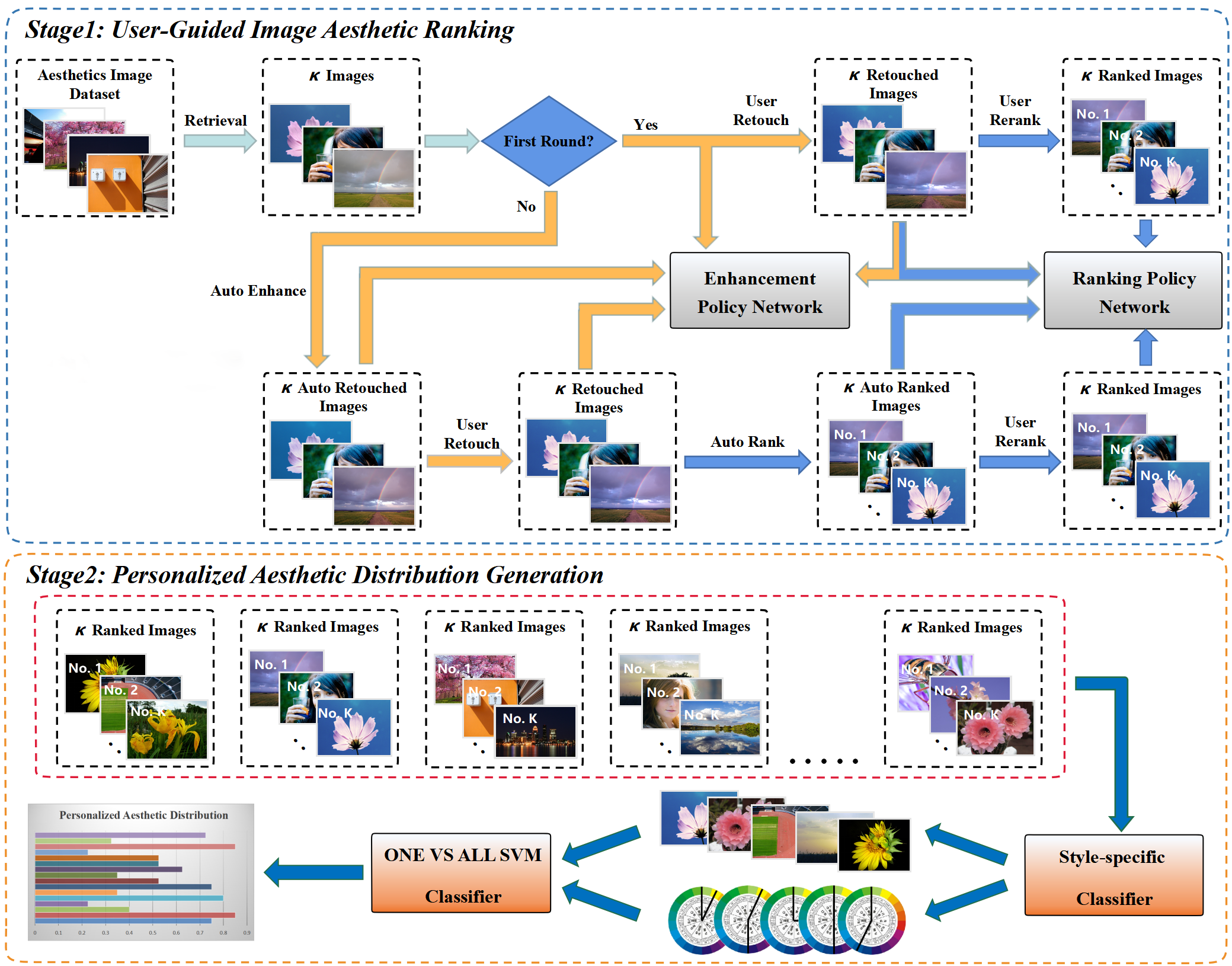}
\caption{
The overview framework of user-guided personalized image aesthetic assessment, which consists of two stages. For Stage $1$, in the process of user interaction, $k$ images are firstly retrieved from the aesthetic database. Then, it is necessary to judge whether it is the first round of interaction. If it is, the original images and the retouched images by users will be directly pushed to the enhancement policy network in pairs for model training, and then these images and the ranked images by users are further pushed to the ranking policy network for model training. Otherwise, the original images are firstly auto-retouched by the enhancement policy network and then retouched manually by users. These two types of retouched images will also be pushed to the enhancement policy network for its fine-tuning. The similar process will be implemented for the ranking policy network. After several iterative operations, these two policy networks are able to output the ranking sequence of personalized enhanced images which satisfy users' aesthetic preferences. For Stage $2$, the aesthetic distribution of users can be generated by multi-label classification of the aesthetic attributes of those modified images, which are from several rounds of interactions.}
\label{fig:framework1}
\vspace{-0.30cm}
\end{figure*}

\section{Framework}
In this paper, we propose a novel user-guided personalized image aesthetic assessment framework based on deep reinforcement learning (illustrated in Fig.~\ref{fig:framework1}). It comprises of two stages: user-guided image aesthetic ranking and personalized aesthetic distribution generation. The framework uses the images in public aesthetic dataset to initialize the policy network, and then uses the user interaction to optimize the personalized policy network. The user-guided image aesthetic ranking is divided into two kinds of situations. If it is the first round of user interaction, the enhancement policy network uses the retrieved images and the user-retouched images for model learning. Subsequently, the retouched images and the user-ranked images are used by the ranking policy network for model learning. Under another situation, the enhancement policy network is used to automatically retouch the retrieved images, and then the ranking policy network is used to automatically rank the retouched images. After several rounds of iterative interaction, we will obtain the final personalized policy network that satisfies the aesthetic preferences of users. Finally, the personalized enhanced image sequences are obtained through the above process, and then multiple labels of images are obtained by aesthetic attribute classifier, and personalized aesthetic distribution is generated based on the ranking of the images and multiple aesthetic attributes, representing the user's aesthetic preferences accurately. In this section, we will describe in detail the procedures and techniques involved above.

\subsection{Interactive Personalized Aesthetic Enhancement}
In order to reduce the influence of the instability and randomness caused by small amount of samples during the training phase, USAR \cite{lv2018usar} introduced the user interaction to optimize the results generated by the primary personalized ranking module. However, USAR did not allow users to retouch images according to personal preference. Different from USAR, this paper improves the interactive strategy and proposes a new module of personalized image aesthetic enhancement.
\begin{figure}[t]
\centering
\vspace{0.1cm}
\includegraphics[width=0.34\textwidth]{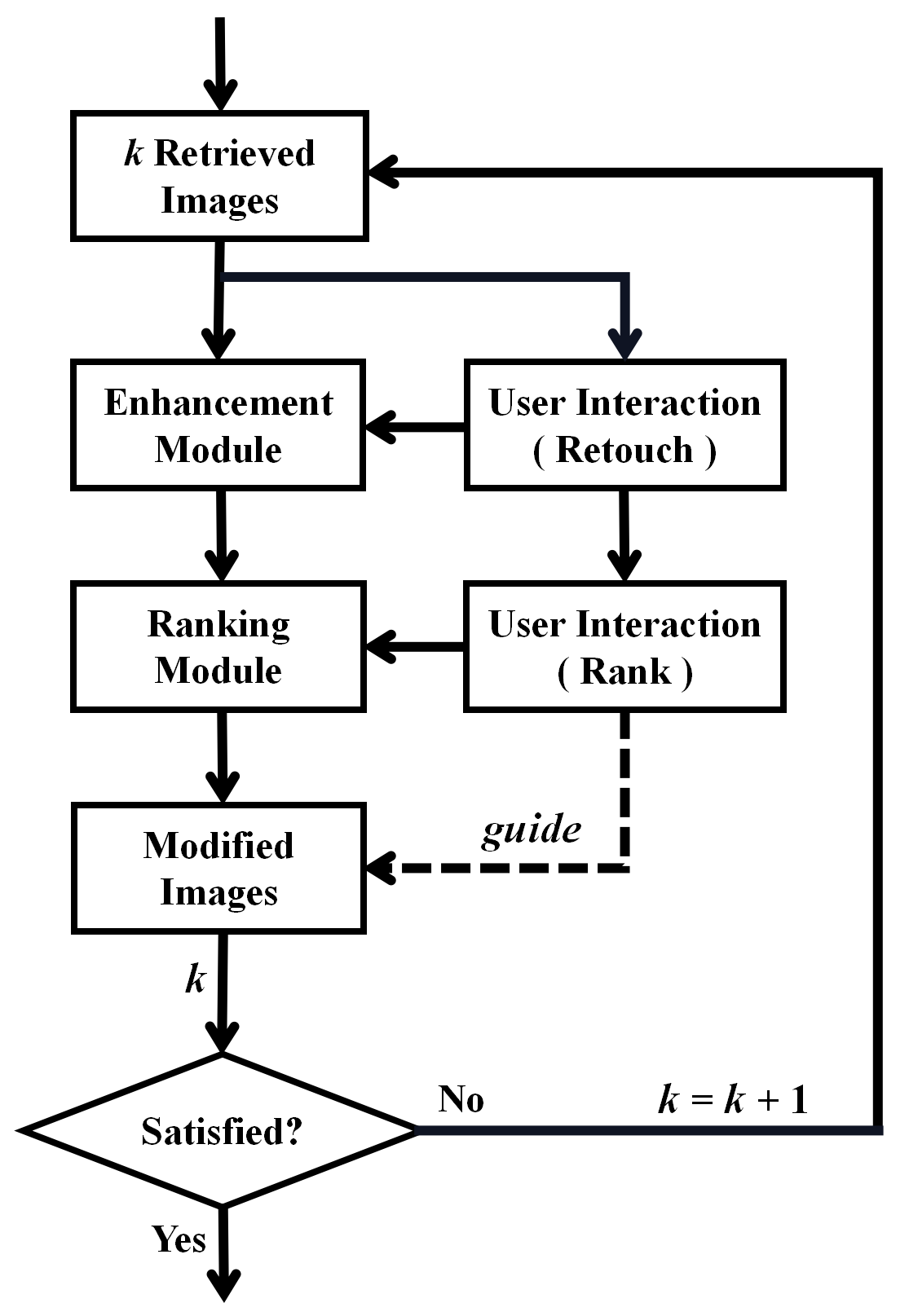}
\caption{The flow chart of user interaction. The desired ranking results will be obtained after $N$ iterations. In each interaction, the user can perform two operations: rank and retouch, which will guide the optimization of the policy networks.}
\label{fig:32UI}
\vspace{-0.5cm}
\end{figure}
\\\indent The process of user interaction is described in detail as follows (shown in Fig.~\ref{fig:32UI}). Our approach generates the personalized aesthetic distribution representing the user's preference through multiple rounds of retouching and ranking interactions of the user. As USAR pointed out, it is clear that only one round of iteration is not so convincing. Therefore, in order to obtain satisfactory results for users, $N$ rounds of adjustment and editing operations will be performed to continuously update the personalized dataset of the user's aesthetic preference. In this paper, the user interaction contains two types of operations: images ranking and images retouching (exposure, gamma, white balance, contrast, tone, color curve, saturation, WNB, \emph{etc.}).  For the retouching process, users can adjust the aesthetic parameters of images to conform to their aesthetic preferences. By recording the retouching process of one user and training the enhancement module, we can acquire more precise personalized aesthetic model.
\\\indent In detail, we describe the image enhancement process as selecting the next operation based on the effect of the current image, and finally simulating the entire sequence of operations for image modification. This process can be regarded as a sequential decision problem. Inspired by Hu\cite{hu2018exposure}, we use Actor-Critic strategy to learn the personalized aesthetic enhancement model. The Actor-Critic strategy is shown in Fig.~\ref{fig:33RL}. The Actor chooses actions according to the strategy. The Critic network updates the value function according to the reward that is given by the environment and guides the actor network to choose better strategy. As the number of iterations increases, the actor obtain the reasonable probability of each action and the critic constantly improve the reward value of action under each state.
\\\indent The above process can be described as $P = \left( {S,A} \right)$. $S$ is state space, and $A$ is action space in the reinforcement learning framework. The intermediate images are generated in the retouching process. $A$ is the set of all filter operations. As we know in \cite{hu2018exposure}, the filter parameters need to be determined after choosing the filter operation when retouching an image. Action space consists of two parts: the discrete set of filter $a_1$ and the continuous set of filter parameter $a_2$. $a_1$ contains eight kinds of filter operations that reflect image aesthetic features, and $a_2$ has different continuous values according to each filter.
\begin{equation}
{a_1} = \left\{ \begin{array}{l}
gamma,{\mathop{\rm exposure}\nolimits} ,contrast,tone,color\\
curve,white\;balance,saturation,W\!\!N\!\!B
\end{array} \right\}
\end{equation}
\\\indent Therefore, the strategy of actor selection also consists of two parts: $\pi {\rm{ = }}({\pi _1},{\pi _2})$ , $\pi_1$ represents the probability distribution for each filter, and $\pi_2$ is the filter parameters which are generated by the actor network. The Critic network is in charge of evaluating the result.
\begin{figure}[t]
\centering
\includegraphics[width=0.34\textwidth]{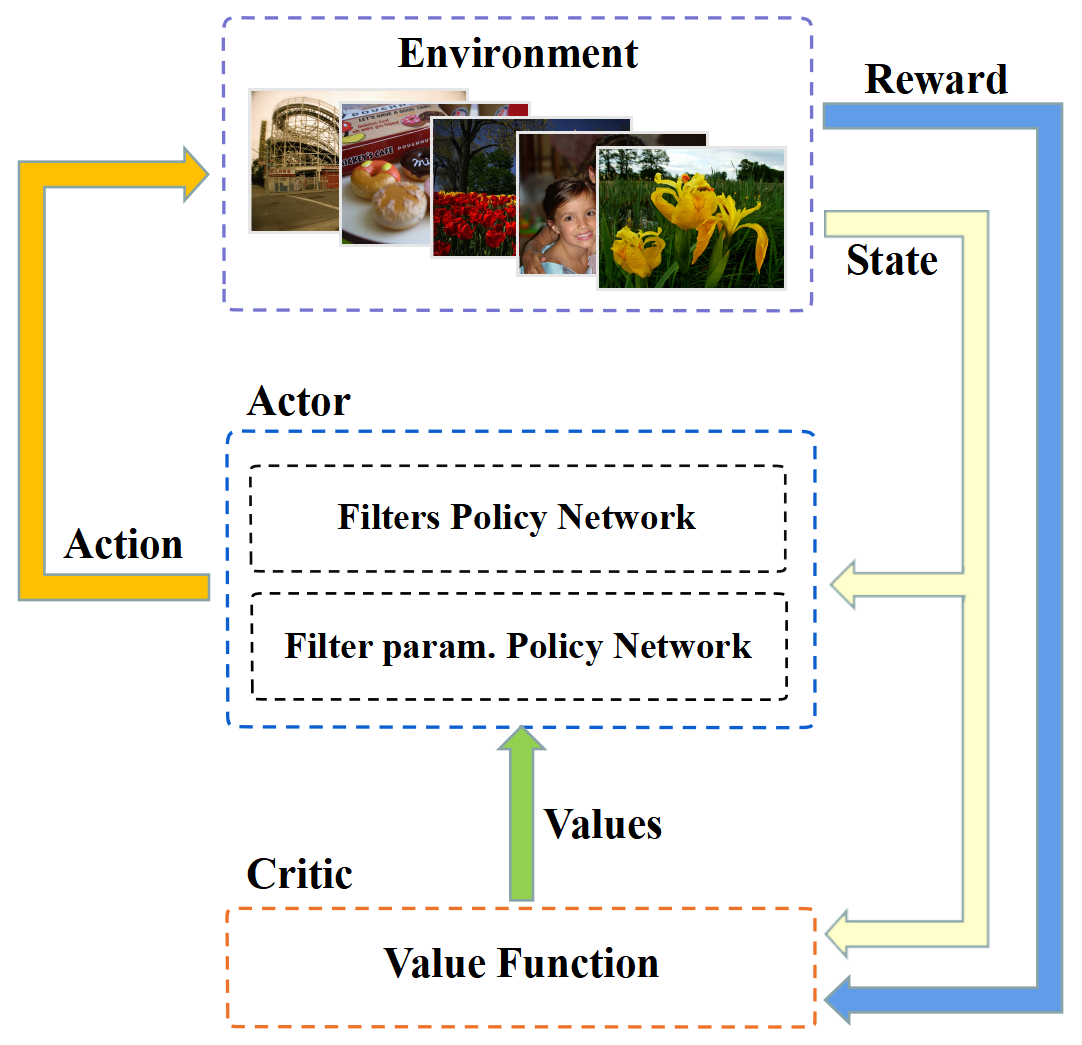}
\caption{Deep reinforcement image aesthetic enhancement. The Actor comprises of filters policy network and filter parameter policy network. The Critic belongs to the value-based learning algorithm. The former requires reward and punishment information to adjust the probability of taking various filters and filter parameters in different states. The latter will be updated in a single step to calculate the reward and punishment value for each step.}
\label{fig:33RL}
\vspace{-0.2cm}
\end{figure}
\begin{figure}[t]
\centering
\includegraphics[width=0.46\textwidth]{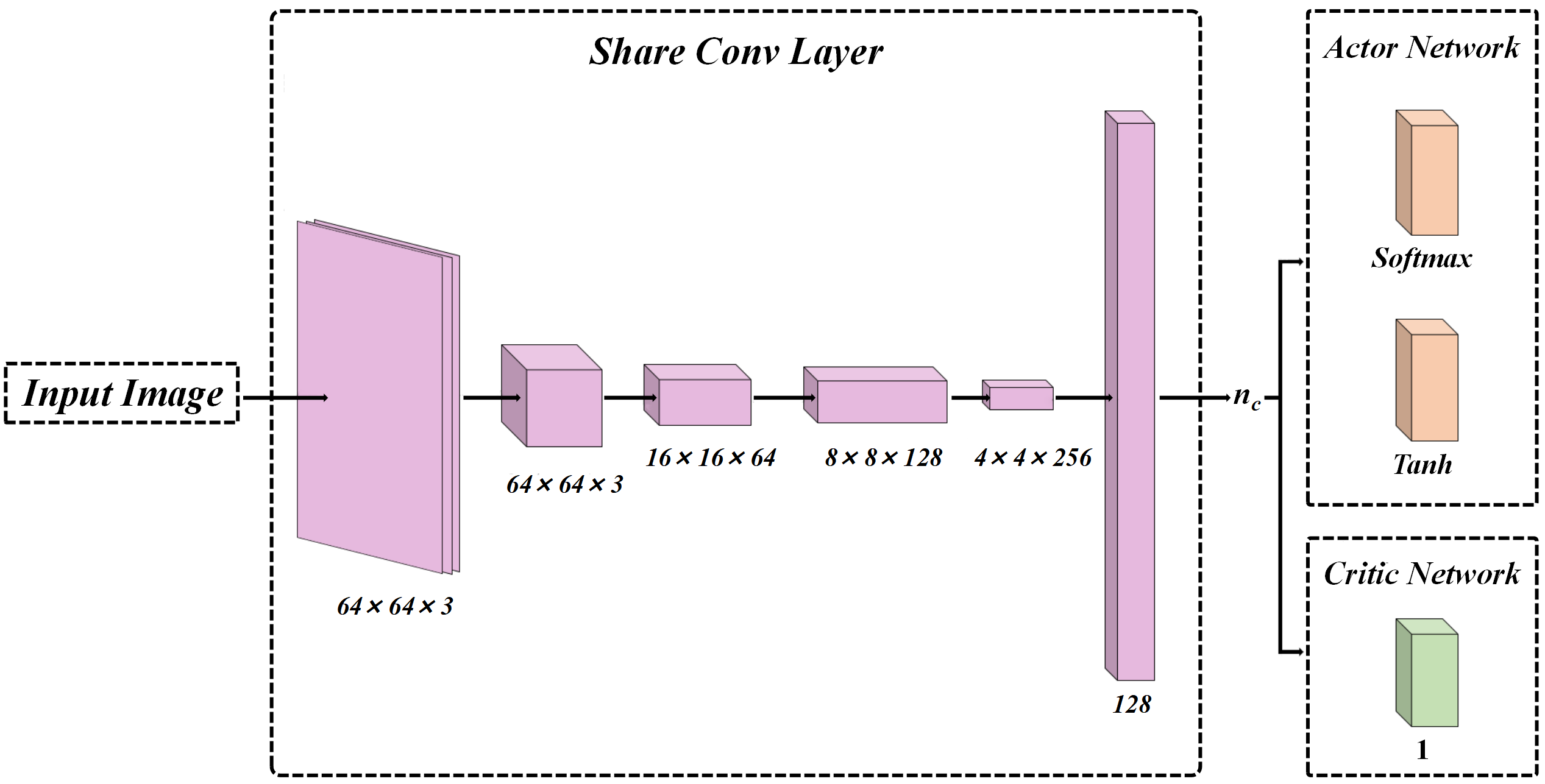}
\caption{The architecture of Actor/Critic network. The Actor network shares the same convolution layer as the Critic network, and produces different results through different output layers $n_c$. Two strategies will be generated in the Actor network, one of which is followed by $Softmax$ to randomly generate discrete filter ID, and the other is followed by $Tanh$ to calculate the continuous parameters of the filter. The Critic network is followed by a fully connected layer to produce a value.}
\label{fig:ACNetwork}
\vspace{-0.5cm}
\end{figure}
\\\indent Our actor-critic network consists of 5 convolutional layers and one fully connected layer (shown in Fig.~\ref{fig:ACNetwork}). The expected value of critic network is:
\begin{equation}
{V_\pi }\left( s \right) = {E_\pi }\left[ {r + \gamma {V_\pi }\left( {s'} \right)} \right]
\end{equation}
\\\indent In order to evaluate the strategy, the value function is defined as follows:
\begin{equation}
{Q_\pi }(s,a) = R _s^a + \gamma {V_\pi }(s')
\end{equation}
\\\indent The loss function in the critic network is defined as follows:
\begin{equation}
{L_c} = \frac{1}{n}\sum\limits_{i = 1}^n {{{[{A_\pi }(s,a)]}^2}}
\end{equation}
where ${A_\pi }(s,a)$ is the Monte Carlo estimation of the optimization function.
\begin{equation}
{A_\pi }(s,a) = {Q_\pi }(s,a) - {V_\pi }(s) = r + \gamma {V_\pi }(s') - {V_\pi }(s)
\end{equation}
\\\indent Actor network chooses actions based on strategy $\pi$ to obtain more return values. The objective of actor network is to maximize the $L_a$.
\\\indent Different from Hu's work \cite{hu2018exposure},  we focus on personalized aesthetic enhancement for different users, while introducing user interaction to achieve accurate extraction of user aesthetic preferences. Therefore, in order to use small amount of user interaction to obtain satisfactory image retouching results, we adopt the following strategy: the enhancement policy network first trains a generic aesthetic enhancement model offline by extracting generic aesthetic features on the publicly available dataset, and then online fine-tunes this model using the images modified by user interaction.

\subsection{Interactive Personalized Aesthetic Ranking}
Recently, many approaches are proposed to solve the problem of image ranking. The personalized image aesthetic ranking becomes even more challenging due to three following reasons. \emph{1)} Due to the high abstraction of image aesthetic features and user's aesthetic preferences, the approach is required to have strong ability of aesthetic feature extraction. \emph{2)} The existing public dataset highlights the popular aesthetics, while there are some differences with the individual aesthetic preferences of different users. \emph{3)} For personalized image aesthetic ranking, the approach must be able to quickly learn different aesthetic preferences of different users with relatively small amount of data.
\\\indent To address these challenges, we introduce the Deep Q-Network (DQN) \cite{mnih2015human}  to realize personalized image aesthetic ranking. Specifically, we represent the sequence of images as continuous state features, the movement of the image as continuous action features, and use them as the input to DQN to predict the potential reward (i.e., whether close to the real ranking by the user). At the same time, our approach can leverage large amounts of public aesthetic data and then quickly adapt to the different aesthetic preferences of different users by online fine-tuning DQN. Meanwhile, DQN is different from other machine learning methods, it needs to constantly take action according to the feedback. Thus we propose to combine user interaction (i.e., user reranks the images in the image sequence)  as feedback.
\\\indent The DQN models the probability that a user may move a particular image in an image sequence and how to move it. In the setting of DQN, the similarity of image sequences and that provided by the user is essentially the reward available to the agent. Therefore, we can model the total reward as Equation \ref{equation:R1}.
\begin{equation}
R_{s,a}\;{\rm{ = }}\; Q(s,a)\;{\rm{ = }}\; r_{immediate}+ {\alpha}r_{future}
\label{equation:R1}
\end{equation}
where state $s$ is represented by the current image sequence, action $a$ is represented by the position and the direction of movement of the image, $r_{immediate}$ represents the reward of the current situation (i.e., the similarity between the current image sequence state and the final image sequence state), and $r_{future}$ represents the estimated future rewards of the agent. $\alpha$ is a discount factor to balance the relative importance of immediate reward and future reward of action $a$. Specifically, given $s$ as the current state, we use the strategy of Double DQN\cite{van2016deep}  to predict the future total reward by taking action $a$ at time $t$ as in Equation \ref{equation:R2}.
\begin{equation}
\begin{split}
R_{s,a,t}&\;{\rm{ = }}\; r_{a,t+1}\\
&\;+\;{\alpha}Q(s_{a,t+1},\arg{\max{Q(s_{a,t+1},a'; {\theta}_t);{\theta}'_t))}}
\end{split}
\label{equation:R2}
\end{equation}
\begin{figure}[t]
\centering
\includegraphics[width=0.8\linewidth]{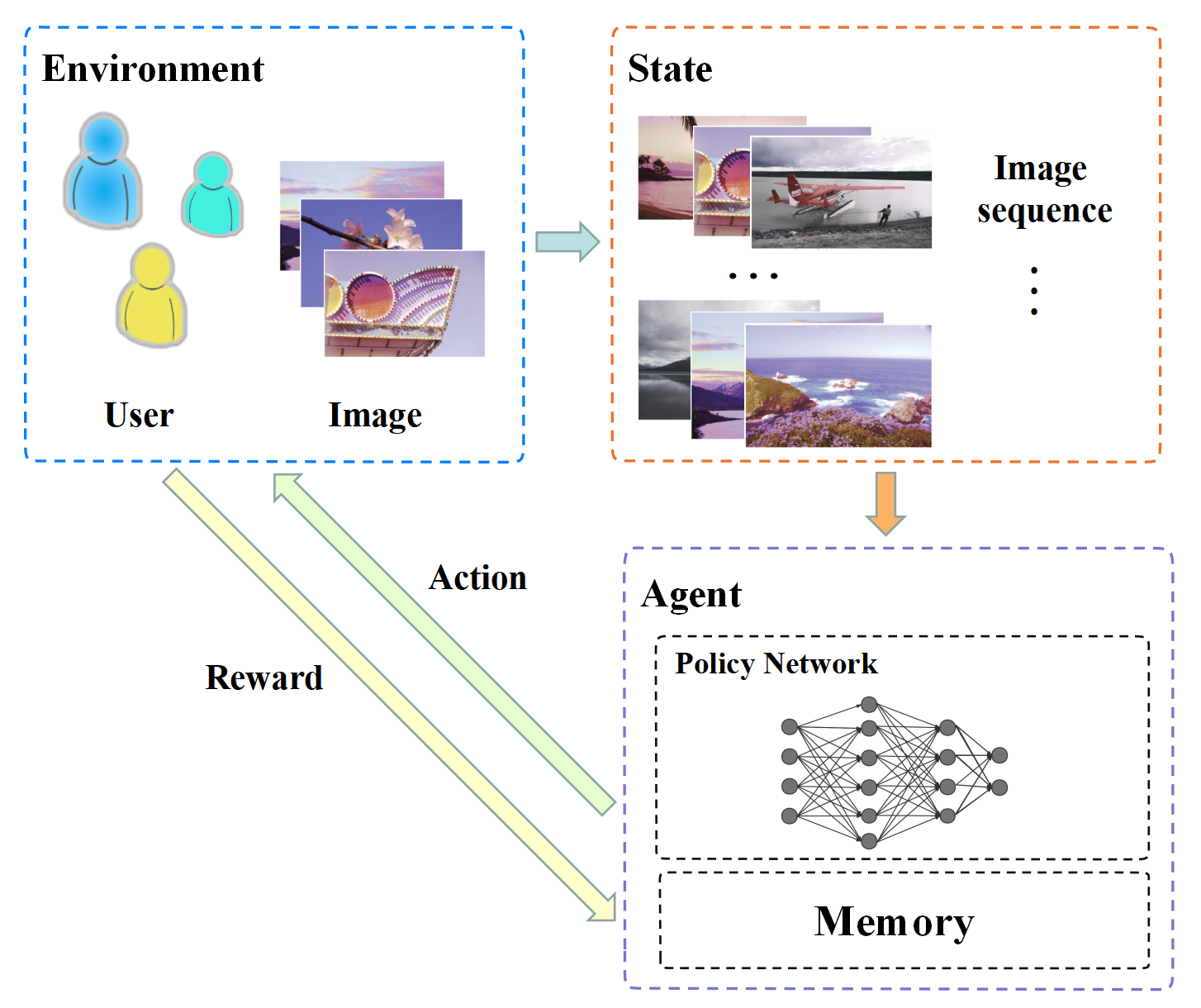}
\caption{Deep reinforcement image aesthetic ranking. At time $t$, the agent receives the state $s_t$ of image sequence from the environment. The agent uses its policy network to choose an appropriate action $a$. Once $a$ is executed, the environment is transferred to the next step, and the next status $s_{t+1}$ of image sequence is provided as well as the reward $r_{t+1}$. At the same time, the image sequence and feedback will be stored in memory. The agent uses knowledge ($s_t$,$a_t$,$s_{t+1}$,$a_{t+1}$,$r_{t+1}$) in the memory to learn and improve its policy network.}
\label{fig:rank}
\vspace{-0.3cm}
\end{figure}
where $r_{a,t+1}$ represents the immediate reward by taking action $a$. $\theta_t$ and ${\theta}'_t$ are two different sets of parameters of the policy network. In this equation, when action $a$ is selected, the agent will speculate the next state $s_{a,t+1}$. Moreover, the action $a'$ that gives the maximum future reward is selected on the next state $s_{a,t+1}$ based on $\theta_t$. After this, the estimated future reward for the given state $s_{a,t+1}$ is calculated based on ${\theta}'_t$. For each fixed number of iterations, $\theta_t$ will replace ${\theta}'_t$. This strategy has been proved to eliminate overly optimistic estimates of Q\cite{van2016deep}. Based on this, our agent will be able to make decisions considering both immediate and future situations.

The deep reinforcement aesthetic ranking can be shown as Fig.~\ref{fig:rank}. In this part, the pool of users and images constitutes the environment, and ranking policy network play the role of agent. The state is defined as the representation of the image sequence and action is defined as the position and direction of movement for images. At each interaction, the state (i.e., images sequence) is pushed to the agent. The agent will select the best action (i.e., the position and direction of movement) and fetch interactive feedback as the reward. Specifically, the reward is composed of the similarity of the image sequence. All these image sequences and feedback log will be stored in the memory of the agent. After certain fixed time, the agent will extract the experience from memory to update its ranking policy network.

\subsection{Personalized Aesthetic Distribution Generation}
After obtaining the ranking sequence of personalized retouched images, we turn to generate the personalized aesthetic distribution to match the user's aesthetic preferences. However, the personalized image aesthetic ranking mainly extracts deep implicit features from user-specific images and does not highlight why users show these specific aesthetic preferences. Therefore, we introduce the image aesthetic attribute classifier to complete the image aesthetic classification. Furthermore, inspired by the multi-label classification task, we generate the user's personalized aesthetic distribution based on the ranking sequence of personalized retouched images. The aesthetic attributes are used in this paper listed in Table \ref{table:AA}.

Once labeled by the well-designed aesthetic attribute classifier, images with multiple aesthetic attribute labels are pushed to our pre-trained one vs all classifier to generate personalized aesthetic distributions. The personalized aesthetic distribution is calculated by concatenating the specific distributions of all single styles as follows:
\begin{equation}
{D_{ugiar}}{\rm{ = }}\sum\limits_{i = 1}^c {\frac{{{r_c} - i + 1}}{{{r_c}\sum c }}} {a_i}
\end{equation}
where $D_{ugiar}$ indicates the personalized distribution of user-specific images, $r_c$ is the ranking of $c$\emph{-th} image's distribution. And $a_i$ represents the aesthetic score of $i$\emph{-th} image.


Given new testing images, we need to do the following two tasks at the same time to evaluate them: \emph{1)} We push these testing images to the aesthetic attribute classifier to generate test distribution $D_{test}$. Then the prediction score is calculated by the correlation $S$ between $D_{ugiar}$ and $D_{test}$ with Equation~\ref{equation:S}. \emph{2)} We send them to users for watching and assessment, and use the generated scores as the ground truth. Finally, we can get the final result by comparing the prediction and the ground truth of users.

\begin{equation}
\begin{small}
S = \frac{
        {
        ({D_{ugiar}} - \overline {{D_{ugiar}} } )
        \sum\limits_{i = 1}^n { ( {D_{test}} - \overline {{D_{test}}} ) }
        }
    }
    {
        {\sqrt {
                    { {  (  {D_{ugiar}} - \overline {  {D_{ugiar}}  }  )  }^2  }
                }
        \sqrt {  \sum\limits_{i = 1}^n  {{   {(   {D_{test}} - \overline {{D_{test}}}  )}^2    }}   }
        }
    }
\end{small}
\end{equation}
\label{equation:S}
\\\noindent where $D_{test}$ represents the distribution of test images, $S$ represents the relative score between $D_{ugiar}$ and $D_{test}$.

\begin{table}[t]\footnotesize
\caption{The chosen aesthetic attribute}
\centering
\begin{tabular}{p{2.8cm}<{\centering}|p{0.8cm}<{\centering}|p{2.6cm}<{\centering}|p{0.8cm}<{\centering}}
\hline
\textbf{Aesthetic attribute} & \textbf{Method}  & \textbf{Aesthetic attribute} & \textbf{Method} \\
\hline
Rule of Thirds &  \cite{mavridaki2015comprehensive} & Tone & \cite{dhar2011high} \\
\hline
Center Composition & \cite{mavridaki2015comprehensive} & Use of Light & \cite{tang2013content} \\
\hline
HROT & \cite{mavridaki2015comprehensive} & Saturation & \cite{datta2006studying} \\
\hline
Sharpness & \cite{mavridaki2015comprehensive} & Image Size & \cite{park2017personalized} \\
\hline
Pattern & \cite{mavridaki2015comprehensive} & Edge Composition & \cite{park2017personalized} \\
\hline
Complementary Colors & \cite{tang2013content} & Global Texture & \cite{lo2012assessment} \\
\hline
Subordinate Colors & \cite{tang2013content} & SDE & \cite{ke2006design} \\
\hline
Cooperate Colors & \cite{tang2013content} & Hue Count & \cite{ke2006design} \\
\hline
Complexity feature & \cite{tang2013content} & Depth of Field & \cite{dhar2011high} \\
\hline
\end{tabular}
\label{table:AA}
\end{table}

\section{Experiments}
In this section, we extensively evaluate the performance of the proposed UG-PIAA  on the two public datasets and compare them with other state-of-the-art methods.
\subsection{Dataset}
The AVA dataset is a large dataset collected by \cite{murray2012ava} for image aesthetic visual analysis, which contains over 255,000 images and is rated by amateur photographers based on image aesthetic qualities. Each image is scored by an average of 200 people in response to photography contests and is associated with a single challenge theme, with nearly 900 different contests in the AVA. The image ratings range from 1 to 10.
\\\indent
The FLICKR-AES dataset is downloaded from Flickr \cite{2017Flickr} and consists of 40,000 images whose aesthetic levels are marked by AMT. The aesthetic levels range from the minimum of 1 to the maximum of 5 to reflect different levels of image aesthetics. Each image in this dataset is evaluated by five AMT workers who participated in the FLICKR-AES annotation process.
\\\indent It should be noticed that the previous datasets are mainly suitable for generic image aesthetic assessment and enhancement. In contrast, our approach mainly focuses on considering individual preference and learns the image aesthetic features of different users. Due to the high cost of data annotation and privacy, previous datasets have too strict limitations to solve the personalized problem. To obtain the personalized aesthetic assessment, users need to retouch the images to produce the enhancement images according to their aesthetic preference. Considering that the high cost of collecting large-scale personalized datasets and most of common people are not good at using the professional retouching software, we provide a reliable online website to collect the personalized dataset of multiple users with different ages, genders, educational backgrounds, $etc$. Users can use this platform to generate datasets that match their own aesthetic preferences, which can be referred to at https://zzuciisr.github.io/UG$\li$PIAA.
\\\indent Users can choose the images (unlike) to retouch (maybe like) or download (like) when dealing with the recommended images (from FLICKR-AES or AVA dataset) on the website. Once users choose to retouch the images, they can use different operations to polish the images until they are satisfied with them and then save them. During this stage, to select representative attributes, we investigate the mainstream image retouching software on mobile phones currently on the market and select the top eight operations for most of users. After the retouched operations, these images will be used as target images to guide the personalized automatic enhancement. In the modification process, we could get two scores, one for the original image and the other for the retouched image. Users can score the image from 1 and 10. In order to make this score more credible, when the process of personalized collection ends, users are allowed to score images in the second round based on the retouched images. If the images have large score deviations, they will be considered as invalid data. Finally, we rank these images and use them to accomplish our experiment.

\subsection{Experimental setting}
\subsubsection{User recruitment}
We select 60 volunteers from a variety of professions, aged 20 to 45, to obtain user-friendly dataset. Our system will automatically deliver images to volunteers from the dataset. Volunteers are asked to score the images and retouch images based on their preferences. In contrast to USAR \cite{lv2018usar}, our model needs volunteers to retouch and rank the images. The retouched and ranked images are pushed sequentially to the enhancement and ranking policy network in pairs. Once the images are pushed to the policy network, the personalized automatic image enhancement policy network will be trained. Subsequently, The enhanced image is used as the input of the ranking policy network, and the ranking policy network is trained to finally generate a sequence of the ranked images. This process requires multiple rounds of iterations.
\subsubsection{Comparison metric}
To compare the existing GIAA and PIAA methods, we compare the ranking correlation measured by Spearman's $\rho$ between the predicted aesthetics scores and the ground-truth aesthetics scores on the AVA dataset. Suppose $r_k$ represents the $k$\emph{-th} ranking with the score $S_k$, and ${\hat r_k}$ represents the ranking of the user according to the score ${\hat S_k}$. Subsequently, ${d_k} = {r_k} - {\hat r_k}$ is then substituted to:
\begin{equation}
\rho {\rm{ = }}1{\rm{ - }}\frac{{6\times \sum d_k^{2} }}{n\times ({n^{2} - 1})}
\end{equation}
$\rho$ measures the difference between the two rankings. $n$ represents the number of images. The coefficient $\rho$  lies in the range $[\:-1,1\:]$.
\subsubsection{Our approach with different settings}
In order to study the effect of the image enhancement module on the distribution of personalized aesthetics, we set up two sets of comparison experiments with or without image enhancement.

\emph{Ranking without enhancement.} Our UG-PIAA directly uses the public datasets for ranking, and then uses the ranked images to generate aesthetic distributions. Finally, we use the aesthetic distribution of images and ground truth to calculate the correlation between them.

\emph{Ranking with enhancement.}  Since the personalized aesthetic features of the images can be improved by the automatic image enhancer, more reliable personalized distribution of customized enhancement images will be received. Meanwhile, we construct a personalized image collection platform for image aesthetic enhancement and image aesthetic assessment. All workers are asked to rank and retouch the test images according to their preferences. For the generation of benchmark distribution, users need to label the aesthetic attributes listed in Table \ref{table:AA}.

\subsection{Automatically personalized image enhancement}
In this section, we will discuss the results of automatic personalized image enhancement.
\subsubsection{Comparison metric}
To verify automatic enhancement images based on reinforcement learning are more in line with users' aesthetic preferences, SSIM is used to measure the correlation of the outputs of user-friendly enhancer and the ground truth of user's preference.
\begin{table}[t]\footnotesize
\caption{The results of enhancer}
\centering
\begin{tabular}{p{4.0cm}<{\centering}|p{1.6cm}<{\centering}|p{1.4cm}<{\centering}}
\hline
\textbf{Enhancer} & & \textbf{SSIM}   \\
\hline
\multirow{3}*{DPED - Ground truth  \cite{ignatov2017dslr}}& Blackberry & 0.5327 \\
\cline{2-3}
~ & Iphone & 0.4889 \\
\cline{2-3}
~ & Sony & 0.5297 \\
\hline
HE - Ground truth \cite{abdullah2007dynamic}& - & 0.3303 \\
\hline
DHE - Ground truth \cite{abdullah2007dynamic}& - & 0.5344 \\
\hline
Ying - Ground truth \cite{ying2017new}& - & 0.5124 \\
\hline
FEQE - Ground truth \cite{Thang2018Fast}& - & 0.6111 \\
\hline
AGCCWD - Ground truth \cite{AGCCWD}& - & 0.6658 \\
\hline
Son \emph{et al.} \cite{son2019naturalness}& - & 0.5068 \\
\hline
Ours(30-5) - Ground truth & - & \textbf{0.6729} \\
\hline
Users & - &1 \\
\hline
\end{tabular}
\label{table:Enhancer}
\end{table}
\subsubsection{Different settings}
To thoroughly study and compare the automatic personalized image enhancement method in detail, several different settings are implemented as follows and compared with other state-of-the-art methods.
\\\indent To study the impact of the few-shot problem involved in the performance of personalized image aesthetic enhancement, we select 5, 10, 15, 20, 30, and 40 images as the input data to fine-tune the well pre-trained model and called it as ours($i- $), where $i$ = 5, 10, 15, 20, 30, 40. Consequent operation depends on the current visual effect of the images during users retouch an image. Hence, different filter combinations are applied to simulate the user's retouching image sequence. The number of filters ranges from 3-7, which is written as ours($ -j$), where $j$ = 3, 4, 5, 6, 7.

\subsubsection{Results and discussion}
In order to enhance the images, most of the existing methods that automatically beautify images usually retouch the images by using a black-box image enhancer or aesthetic rules. In contrast to those methods, we take into account the user's aesthetic preferences to achieve personalized enhanced images. In this section, we compare our results with the HE(histogram equalization)\cite{abdullah2007dynamic}, DHE(dynamic histogram equalization)\cite{abdullah2007dynamic}, DPED\cite{ignatov2017dslr}, Abdullah \emph{et al.}\cite{abdullah2007dynamic}, Ying \emph{et al.}\cite{ying2017new}, FEQE\cite{Thang2018Fast}, AGCCWD\cite{AGCCWD}, and Son \emph{et al.} \cite{son2019naturalness} to verity that our results are more in line with the users' aesthetic preferences. Experiments show that our method works best in Table \ref{table:Enhancer}.
\\\indent To further explore the effect of the filters on the personalized aesthetic enhancement module proposed in this paper, we experimented the performance of the module by limiting the number of filters and that of input images, and finally used these experimental results to select appropriate numbers. In Table \ref{table:fliter}, it can be seen that when $i$ = 30 and $j$ = 5, the model describes the user's preferences most closely (SSIM is 0.6729). Although the number of filters has been increased, the performance of the model has not been promoted. The experimental results are not only related to the number of filters but also related to the number of images. These two factors need to be coordinated to make the experiment work perfectly. Therefore we also pay attention to the number $i$ of input data, it can be found that the model outperforms best when $i$ = 30. As the number of input increases, more user aesthetic preference information can be obtained. Therefore, the performance on $i$ =30 outperforms that on $i$ = 5, 10, 15, 20. However, psychological research \cite{aestheticfatigue} shows that, along with the increase of number of input, the data quality will decrease because of the individual's aesthetic fatigue. The performance on $i$ = 40 is generally worse than that on $i$ = 30. According to these results, we set $i$ = 30, $j$ = 5 and compare our results with others on the  AVA dataset as detailed in Section~\ref{pirr}.
\begin{table}[t]\footnotesize
\caption{The results of enhancer under different $i$ and $j$}
\centering
\begin{tabular}{p{3.2cm}<{\centering}|p{2.4cm}<{\centering}|p{1.5cm}<{\centering}}
\hline
\textbf{different input number($i$)} & \textbf{different filters($j$)} & \textbf{SSIM}   \\
\hline
\multirow{5}*{Ours(5-$j$)} & Ours(5-3)& 0.5908 \\
\cline{2-3}
~ & Ours(5-4) & 0.6311 \\
\cline{2-3}
~ & Ours(5-5) & 0.6727 \\
\cline{2-3}
~ & Ours(5-6) & 0.6642 \\
\cline{2-3}
~ & Ours(5-7) & 0.6537 \\
\hline

\multirow{5}*{Ours(10-$j$)} & Ours(10-3)& 0.6198 \\
\cline{2-3}
~ & Ours(10-4) & 0.6344 \\
\cline{2-3}
~ & Ours(10-5) & 0.6592 \\
\cline{2-3}
~ & Ours(10-6) & 0.6212 \\
\cline{2-3}
~ & Ours(10-7) & 0.6338 \\
\hline

\multirow{5}*{Ours(15-$j$)} & Ours(15-3)& 0.6319 \\
\cline{2-3}
~ & Ours(15-4) & 0.6521 \\
\cline{2-3}
~ & Ours(15-5) & 0.6552 \\
\cline{2-3}
~ & Ours(15-6) & 0.6246 \\
\cline{2-3}
~ & Ours(15-7) & 0.6343 \\
\hline

\multirow{5}*{Ours(20-$j$)} & Ours(20-3)& 0.6250 \\
\cline{2-3}
~ & Ours(20-4) & 0.6366 \\
\cline{2-3}
~ & Ours(20-5) & 0.6385 \\
\cline{2-3}
~ & Ours(20-6) & 0.6437 \\
\cline{2-3}
~ & Ours(20-7) & 0.6347 \\
\hline

\multirow{5}*{Ours(30-$j$)} & Ours(30-3)& 0.6151 \\
\cline{2-3}
~ & Ours(30-4) & 0.6604 \\
\cline{2-3}
~ & Ours(30-5) & \textbf{0.6729} \\
\cline{2-3}
~ & Ours(30-6) & 0.6551 \\
\cline{2-3}
~ & Ours(30-7) & 0.5487 \\
\hline

\multirow{5}*{Ours(40-$j$)} & Ours(40-3)& 0.6206 \\
\cline{2-3}
~ & Ours(40-4) & 0.6534 \\
\cline{2-3}
~ & Ours(40-5) & 0.6485 \\
\cline{2-3}
~ & Ours(40-6) & 0.6492 \\
\cline{2-3}
~ & Ours(40-7) & 0.5785 \\
\hline

\end{tabular}
\label{table:fliter}
\end{table}

\subsection{Personalized image ranking results}
\label{pirr}
In this section, in order to verify the effectiveness of personalized image aesthetic ranking learned from user's aesthetic preference, we compare the results with other state-of-the-art personalized aesthetic evaluation methods FPMF\cite{o2014collaborative}, PAM\cite{ren2017personalized}, O-D$_{USAR}$\cite{lv2018usar}, NIMA\cite{talebi2018nima}, PA$\li$IAA\cite{li2020personality}, $etc$. which are proposed in recent years.

\begin{table}[t]\scriptsize
\vspace{-0.3cm}
\caption{The results on the FLICKR-AES dataset}
\centering
\begin{tabular}{p{4.2cm}<{\centering}|p{1.8cm}<{\centering}|p{1.4cm}<{\centering}}
\hline
\textbf{Methods} & \textbf{10images}  & \textbf{100images}    \\
\hline
FPMF(only attribute) \cite{o2014collaborative}& 0.511$\pm$0.004  & 0.516$\pm$0.003   \\
\hline
FPMF(only content) \cite{o2014collaborative}& 0.512$\pm$0.002  & 0.516$\pm$0.010   \\
\hline
FPMF(attribute and content) \cite{o2014collaborative}& 0.513$\pm$0.003  &  0.524$\pm$0.007  \\
\hline
PAM(only attribute) \cite{ren2017personalized }& 0.518$\pm$0.003  &  0.539$\pm$0.013  \\
\hline
PAM(only content) \cite{ren2017personalized }& 0.515$\pm$0.004  & 0.535$\pm$0.017   \\
\hline
PAM(attribute and content) \cite{ren2017personalized }& 0.520$\pm$0.003  & 0.553$\pm$0.012   \\
\hline
USAR$\li$PPR \cite{lv2018usar}&  0.521$\pm$0.002 &  0.544$\pm$0.007  \\
\hline
USAR$\li$PAD \cite{lv2018usar}&  0.520$\pm$0.003 &  0.537$\pm$0.003  \\
\hline
USAR$\li$PPR\&PAD \cite{lv2018usar}&  0.525$\pm$0.004 &  0.552$\pm$0.015  \\
\hline
MT$\li$IAA \cite{li2020personality}&  0.523$\pm$0.004 &  0.582$\pm$0.014  \\
\hline
PA$\li$IAA \cite{li2020personality}&  0.543$\pm$0.003 &  0.639$\pm$0.011  \\
\hline
UG-PIAA(ranking without enhancement) & 0.537$\pm$0.004  &  0.629$\pm$0.007  \\
\hline
\textbf{UG-PIAA(ranking with enhancement)} & \textbf{0.559$\pm$0.002}  &  \textbf{0.660$\pm$0.013}  \\
\hline
\end{tabular}
\label{table:Flickr}
\end{table}

\begin{table}[t]\scriptsize
\caption{PERFORMANCE COMPARISON BETWEEN OUR METHOD AND BASELINE METHODS ON AVA DATABASE}
\centering
\begin{tabular}{p{4.5cm}<{\centering}|p{1.2cm}<{\centering}|p{1.2cm}<{\centering}}
\hline
\textbf{Methods} & \textbf{ACC(\%)}  & \textbf{SROCC}  \\
\hline
AVA handcrafted features\cite{murray2012ava} & 68.0 &  -   \\
RAPID \cite{lu2014rapid}& 74.5 &  -  \\
DMA \cite{lu2015deep}& 75.4 &  -   \\
Wang \emph{et al.} \cite{wang2016multi}& 76.9 &  -  \\
Kao \emph{et al.}  \cite{kao2016hierarchical}& 76.2 &  -  \\
BDN \cite{wang2016brain}& 78.1 &  -  \\
Kao \emph{et al.} \cite{kao2017deep}& 79.1 &  -   \\
Zhang \emph{et al.} \cite{zhang2018visual}& 78.8 &  -  \\
Schwarz \emph{et al.} \cite{schwarz2018will}& 75.8 &  -  \\
Kucer \emph{et al.} \cite{kucer2018leveraging} & 81.9 &  -  \\
ILGNet \cite{jin2019ilgnet} & 82.7 &  -  \\
\hline
AlexNet$\li$FT$\li$Conf \cite{kong2016photo} & 71.5 & 0.481  \\
Reg+Rank+Att \cite{kong2016photo}& 75.5 & 0.545    \\
Reg+Rank+Cont \cite{kong2016photo} & 73.4 & 0.541    \\
Reg+Rank+Att+Cont \cite{kong2016photo} & 77.3 & 0.558    \\
USAR$\li$PPR \cite{lv2018usar} & 72.4 & 0.600  \\
USAR$\li$PAD \cite{lv2018usar} & 77.7 & 0.545  \\
USAR$\li$PPR\&PAD \cite{lv2018usar} & 78.1 & 0.578  \\
\hline
NIMA(VGG16) \cite{talebi2018nima} & 80.6 & 0.592  \\
NIMA(Inception-v2) \cite{talebi2018nima} & 81.5 & 0.612  \\
DenseNet121(asethetics) \cite{li2020personality} & 80.5 & 0.630  \\
Inception-v3(asethetics) \cite{li2020personality} & 80.9 & 0.638   \\
PA$\li$IAA(DenseNet121) \cite{li2020personality} & 82.9 & 0.666  \\
PA$\li$IAA(Inception-v3) \cite{li2020personality} & 83.7 & 0.677  \\
UG-PIAA(ranking without enhancement)& 83.2  &  0.659  \\
\textbf{UG-PIAA(ranking with enhancement)} & \textbf{85.1}  &  \textbf{0.692}  \\
\hline
\end{tabular}
\label{table:Ava}
\end{table}

\subsubsection{Result and discussion on FLICKR-AES dataset}
In this dataset, we compare our approach with that of FPMF \cite{o2014collaborative}, PAM \cite{ren2017personalized}, USAR\cite{lv2018usar}, MT$\li$IAA \cite{li2020personality} and PA$\li$IAA\cite{li2020personality} to verity the effectiveness of the proposed method. The average score and standard deviation are obtained on the FLICKR-AES dataset with the number of 10 and 100 images in Table \ref{table:Flickr}.

\begin{figure*}[t]
\centering
\includegraphics[width=0.96\textwidth]{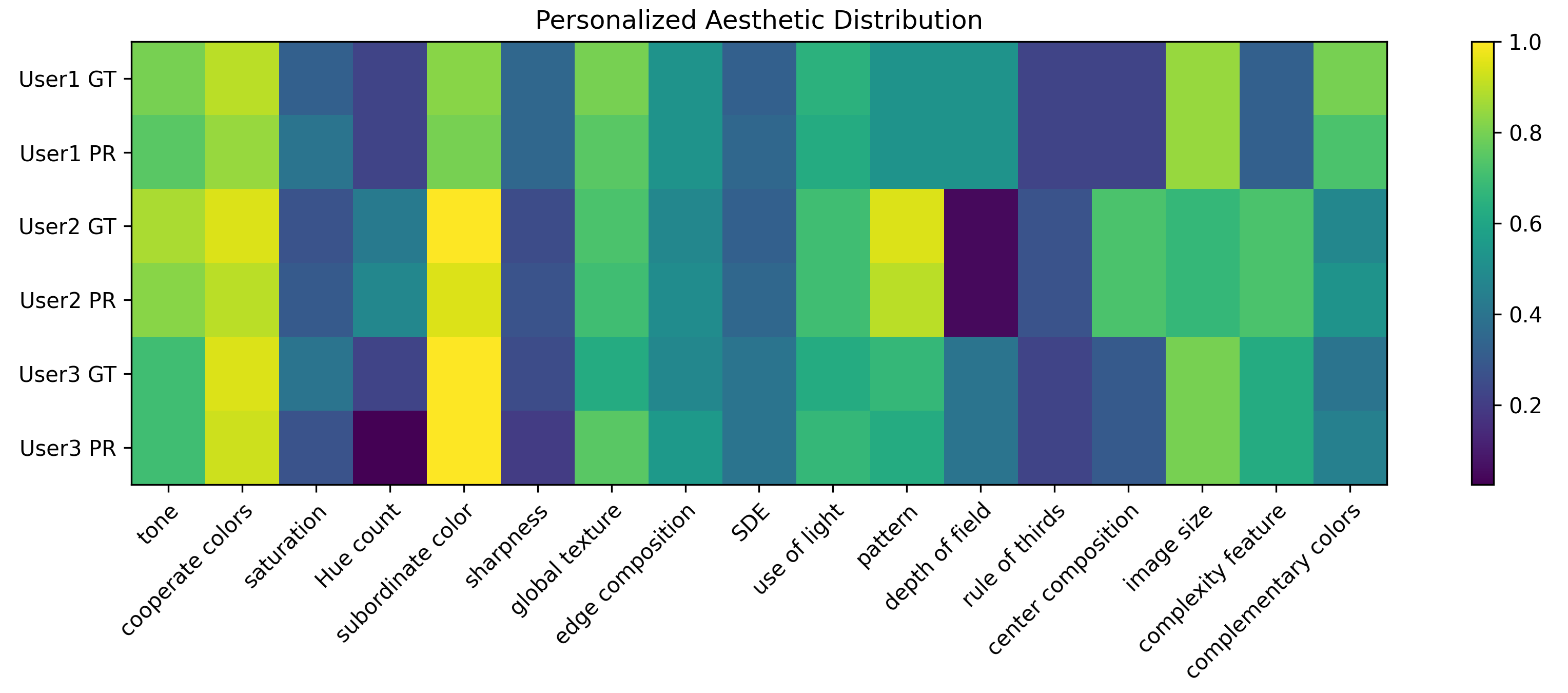}
\caption{The heat map represents the correlation between the ground-truth of image aesthetic distribution and the predicted results of users.} 
\label{fig:USAD}
\end{figure*}

Through qualitative analysis, FPMF evaluates personalized image aesthetic quality through the combination of two different frameworks. PAM solves the problem of personalized aesthetics by predicting the deviation between individual perception and generic image aesthetics. PIAA introduces personality characteristics and meta-learning into the field of personalized aesthetic assessment and optimizes the results of the aesthetic assessment. USAR$\li$PPR\&PAD improves the user's aesthetic preference through a small number of user interactions. PA$\li$IAA introduces personality traits and multi-task deep learning to the personalized image aesthetics assessment. Our approach achieves excellent performance with the addition of personalized image aesthetics enhancement. 
\\\indent The aesthetic distribution is limited if the user can only select satisfactory images. In order to overcome this problem, we encourage users to generate aesthetic attributes that may not exist in the dataset by adding retouched operations in the user interaction stage. On the one hand, the aesthetic distribution can be optimized by adding new aesthetic attributes. On the other hand, it can highlight the user's personalized aesthetic preferences and avoid missing aesthetic attributes due to a small number of interactions. The results show that using the personalized dataset can obtain more accurate personalized aesthetic expression.

\subsubsection{Results and discussion on AVA dataset}
The image ranking correlation and the image classification accuracy on the AVA Dataset are summarized in Table \ref{table:Ava}. By adding the retouched operation in the user interaction stage, the precision of personalized aesthetic distribution in line with user-specific preference will be obtained as shown in Fig.~\ref{fig:USAD}. Then the personalized aesthetic assessment will be more accurate. The setting of ours (ranking with enhancement) outperforms best.

\begin{table}[t]
    \vspace{-0.2cm}
    \caption{The results of ranking correlation under different $t$ and $n$}
    \centering
	\begin{tabular}{ | p{2.6cm}<{\centering} | p{1.4cm}<{\centering} |p{1.4cm} <{\centering}| p{1.4cm}<{\centering} |}
		\hline
		 	&\multicolumn{3}{|c|}{\textbf{\footnotesize{The number of images in interaction ($n$)}}} \\
		\hline
	        \textbf{\footnotesize{Interaction times ($t$)}} 	& 5		& 10		 & 15  \\
		\hline
		1 & 0.5642		& 0.5373		  & 0.5123  \\
		\hline
		2 & 0.6117		&  0.5869		  & 0.5485  \\
		\hline
		3 & 0.6401 		& 0.6191		  & 0.5791  \\
		\hline
		4 & 0.6869		& 0.6521		  & 0.6058  \\
		\hline
		5 & 0.6926 		& 0.6624		 & 0.6478  \\
		\hline
	\end{tabular}
	\vspace{-0.1cm}
	\label{table:mk}
\end{table}

\begin{table}[t]\footnotesize
    \caption{The average time for one interaction}
    \centering
    \begin{tabular}{p{5.0cm}<{\centering}|p{2.6cm}<{\centering}}
    \hline
        \textbf{\footnotesize{The number of images in interaction ($n$)}} &\textbf{ \footnotesize{Average time (min)} }  \\
    \hline
        5 & 1.83   \\
    \hline
        10 & 4.18 \\
    \hline
        15 & 6.81   \\
    \hline

    \end{tabular}
    \label{table:time}
    \vspace{-0.3cm}
\end{table}

We further explore whether our approach can stabilize user preferences by limiting the parameters of $n$ and $t$, where $n$ is the number of images pushed to the user during each interaction and $t$ is the total number of interactions between UG-PIAA and the user. Based on this, we use the data from different users in our established dataset and calculate the Spearman's $\rho$ between the real user ranking and predicted individual ranking using the parameters of variation of $n$ and $t$ in Table \ref{table:mk}.
\\\indent The Table \ref{table:mk} clearly shows that different parameters $n$ and $t$ have a certain influence on the results. At the same time, in order to further analyze how parameters $n$ and $t$ should be set during the interaction process, we change the number of images pushed to users in each interaction process and count the average time spent by different users to complete all the retouching and ranking operations in the interaction process as Table \ref{table:time}.
\\\indent By analyzing the user's feedback of the retouching and ranking operations in the real interaction, we have come to the following conclusion. The impact of the different number of images on the user's retouching time is linear. However, when more images need to be ranked, the time consumed by the user will increase exponentially. During the ranking operation, we notice that the real situation of users is that if the number of images doubles, the user can not watch all images at a glance to get the final ranking result and needs to watch them several times to have a general judgment. Moreover, when the number of images increases, some users need to adjust the order of some images after the first ranking. At the same time, combined with the statistical data in Table \ref{table:mk} and Table \ref{table:time}, we comprehensively analyzed the accuracy of the aesthetic ranking and the time of the interaction process. We concluded that the number of images pushed to the user during each interaction process should not be too much. Through a small number of interactions with enough times, not only better accuracy can be obtained, but also the interaction time can be minimized, and the user's aesthetic fatigue can be reduced.

\section{Conclusion}
In this paper, we propose a novel user-guided personalized image aesthetic assessment framework. Particularly, personalized image aesthetic enhancement and user interaction are introduced into the aesthetic assessment to describe the aesthetic preferences of users, and we explore their positive effects on personalized image aesthetic assessment. Extensive experimental results on the publicly available datasets demonstrate the advancement and effectiveness of the method from two aspects: \emph{1)} The proposed personalized image aesthetic ranking achieves outstanding results compared to the existing approaches. \emph{2)} Our approach achieves excellent performance in predicting the aesthetic preferences of various users, and the generated aesthetic distribution is closer to the real aesthetic distribution of users.

In future work, it would be meaningful: \emph{1)} We use the FLICKR-AES and AVA dataset for training to realize the image aesthetic enhancement and ranking, but there is still large gap compared to the 1.4 $\times$ 10$^7$ images used for image classification in ImageNet. In the future, we will consider how to transfer the knowledge extracted from ImageNet to solve the aesthetic problem or generate another massive datasets of high-quality aesthetics enhancement and assessment images in an automatic way. \emph{2)} Our experiments verify the positive effect of personalized aesthetic enhancement on personalized aesthetic evaluation, and we believe that using more powerful personalized aesthetic enhancement measures (e.g., optimizing the image composition) will achieve better results.


%

\ifCLASSOPTIONcaptionsoff
  \newpage
\fi



%
\bibliographystyle{IEEEtran}
\bibliography{IEEEabrv,xu20191126}

\begin{thebibliography}{10}
\providecommand{\url}[1]{#1}
\csname url@samestyle\endcsname
\providecommand{\newblock}{\relax}
\providecommand{\bibinfo}[2]{#2}
\providecommand{\BIBentrySTDinterwordspacing}{\spaceskip=0pt\relax}
\providecommand{\BIBentryALTinterwordstretchfactor}{4}
\providecommand{\BIBentryALTinterwordspacing}{\spaceskip=\fontdimen2\font plus
\BIBentryALTinterwordstretchfactor\fontdimen3\font minus
  \fontdimen4\font\relax}
\providecommand{\BIBforeignlanguage}[2]{{%
\expandafter\ifx\csname l@#1\endcsname\relax
\typeout{** WARNING: IEEEtran.bst: No hyphenation pattern has been}%
\typeout{** loaded for the language `#1'. Using the pattern for}%
\typeout{** the default language instead.}%
\else
\language=\csname l@#1\endcsname
\fi
#2}}
\providecommand{\BIBdecl}{\relax}
\BIBdecl

\bibitem{tong2004classification}
H.~Tong, M.~Li, H.-J. Zhang, J.~He, and C.~Zhang, ``Classification of digital
  photos taken by photographers or home users,'' in \emph{Pacific-Rim
  Conference on Multimedia}.\hskip 1em plus 0.5em minus 0.4em\relax Springer,
  2004, pp. 198--205.

\bibitem{datta2006studying}
R.~Datta, D.~Joshi, J.~Li, and J.~Z. Wang, ``Studying aesthetics in
  photographic images using a computational approach,'' in \emph{European
  conference on computer vision}.\hskip 1em plus 0.5em minus 0.4em\relax
  Springer, 2006, pp. 288--301.

\bibitem{luo2008photo}
Y.~Luo and X.~Tang, ``Photo and video quality evaluation: Focusing on the
  subject,'' in \emph{European conference on computer vision}.\hskip 1em plus
  0.5em minus 0.4em\relax Springer, 2008, pp. 386--399.

\bibitem{wong2009saliency}
L.-K. Wong and K.-L. Low, ``Saliency-enhanced image aesthetics class
  prediction,'' in \emph{2009 16th IEEE International Conference on Image
  Processing (ICIP)}.\hskip 1em plus 0.5em minus 0.4em\relax IEEE, 2009, pp.
  997--1000.

\bibitem{li2009aesthetic}
C.~Li and T.~Chen, ``Aesthetic visual quality assessment of paintings,''
  \emph{IEEE Journal of selected topics in Signal Processing}, vol.~3, no.~2,
  pp. 236--252, 2009.

\bibitem{nishiyama2011aesthetic}
M.~Nishiyama, T.~Okabe, I.~Sato, and Y.~Sato, ``Aesthetic quality
  classification of photographs based on color harmony,'' in \emph{Proceedings
  of the IEEE conference on computer vision and pattern recognition}.\hskip 1em
  plus 0.5em minus 0.4em\relax IEEE, 2011, pp. 33--40.

\bibitem{dhar2011high}
S.~Dhar, V.~Ordonez, and T.~L. Berg, ``High level describable attributes for
  predicting aesthetics and interestingness,'' in \emph{Proceedings of the IEEE
  conference on computer vision and pattern recognition}.\hskip 1em plus 0.5em
  minus 0.4em\relax IEEE, 2011, pp. 1657--1664.

\bibitem{marchesotti2011assessing}
L.~Marchesotti, F.~Perronnin, D.~Larlus, and G.~Csurka, ``Assessing the
  aesthetic quality of photographs using generic image descriptors,'' in
  \emph{2011 International conference on computer vision}.\hskip 1em plus 0.5em
  minus 0.4em\relax IEEE, 2011, pp. 1784--1791.

\bibitem{tang2013content}
X.~Tang, W.~Luo, and X.~Wang, ``Content-based photo quality assessment,''
  \emph{IEEE Transactions on Multimedia}, vol.~15, no.~8, pp. 1930--1943, 2013.

\bibitem{PIAA2020}
H.~{Zhu}, L.~{Li}, J.~{Wu}, S.~{Zhao}, G.~{Ding}, and G.~{Shi}, ``Personalized
  image aesthetics assessment via meta-learning with bilevel gradient
  optimization,'' \emph{IEEE Transactions on Cybernetics}, 2020.

\bibitem{lu2014rapid}
X.~Lu, Z.~Lin, H.~Jin, J.~Yang, and J.~Z. Wang, ``Rapid: Rating pictorial
  aesthetics using deep learning,'' in \emph{Proceedings of the 22nd ACM
  international conference on Multimedia}.\hskip 1em plus 0.5em minus
  0.4em\relax ACM, 2014, pp. 457--466.

\bibitem{kong2016photo}
S.~Kong, X.~Shen, Z.~Lin, R.~Mech, and C.~Fowlkes, ``Photo aesthetics ranking
  network with attributes and content adaptation,'' in \emph{European
  conference on computer vision}.\hskip 1em plus 0.5em minus 0.4em\relax
  Springer, 2016, pp. 662--679.

\bibitem{lu2015deep}
X.~Lu, Z.~Lin, X.~Shen, R.~Mech, and J.~Z. Wang, ``Deep multi-patch aggregation
  network for image style, aesthetics, and quality estimation,'' in
  \emph{International conference on computer vision}, 2015, pp. 990--998.

\bibitem{jin2018ilgnet}
X.~Jin, L.~Wu, X.~Li, X.~Zhang, J.~Chi, S.~Peng, S.~Ge, G.~Zhao, and S.~Li,
  ``Ilgnet: inception modules with connected local and global features for
  efficient image aesthetic quality classification using domain adaptation,''
  \emph{IET Computer Vision}, vol.~13, no.~2, pp. 206--212, 2018.

\bibitem{jin2016image}
B.~Jin, M.~V.~O. Segovia, and S.~S{\"u}sstrunk, ``Image aesthetic predictors
  based on weighted cnns,'' in \emph{2016 IEEE International Conference on
  Image Processing (ICIP)}.\hskip 1em plus 0.5em minus 0.4em\relax Ieee, 2016,
  pp. 2291--2295.

\bibitem{kang2014convolutional}
L.~Kang, P.~Ye, Y.~Li, and D.~Doermann, ``Convolutional neural networks for
  no-reference image quality assessment,'' in \emph{Proceedings of the IEEE
  conference on computer vision and pattern recognition}, 2014, pp. 1733--1740.

\bibitem{wang2016brain}
Z.~Wang, S.~Chang, F.~Dolcos, D.~Beck, D.~Liu, and T.~S. Huang,
  ``Brain-inspired deep networks for image aesthetics assessment,'' \emph{arXiv
  preprint arXiv:1601.04155}, 2016.

\bibitem{mai2016composition}
L.~Mai, H.~Jin, and F.~Liu, ``Composition-preserving deep photo aesthetics
  assessment,'' in \emph{Proceedings of the IEEE conference on computer vision
  and pattern recognition}, 2016, pp. 497--506.

\bibitem{yeh2010personalized}
C.-H. Yeh, Y.-C. Ho, B.~A. Barsky, and M.~Ouhyoung, ``Personalized photograph
  ranking and selection system,'' in \emph{Proceedings of the 18th ACM
  international conference on Multimedia}.\hskip 1em plus 0.5em minus
  0.4em\relax ACM, 2010, pp. 211--220.

\bibitem{yeh2014personalized}
C.-H. Yeh, B.~A. Barsky, and M.~Ouhyoung, ``Personalized photograph ranking and
  selection system considering positive and negative user feedback,'' \emph{ACM
  Transactions on Multimedia Computing, Communications, and Applications
  (TOMM)}, vol.~10, no.~4, p.~36, 2014.

\bibitem{jin2018predicting}
X.~Jin, L.~Wu, X.~Li, S.~Chen, S.~Peng, J.~Chi, S.~Ge, C.~Song, and G.~Zhao,
  ``Predicting aesthetic score distribution through cumulative jensen-shannon
  divergence,'' in \emph{Thirty-Second AAAI Conference on Artificial
  Intelligence}, 2018.

\bibitem{lv2018usar}
P.~Lv, M.~Wang, Y.~Xu, Z.~Peng, J.~Sun, S.~Su, B.~Zhou, and M.~Xu, ``{USAR:} an
  interactive user-specific aesthetic ranking framework for images,'' in
  \emph{2018 {ACM} Multimedia Conference on Multimedia Conference, {MM} 2018,
  Seoul, Republic of Korea, October 22-26, 2018}, S.~Boll, K.~M. Lee, J.~Luo,
  W.~Zhu, H.~Byun, C.~W. Chen, R.~Lienhart, and T.~Mei, Eds.\hskip 1em plus
  0.5em minus 0.4em\relax {ACM}, 2018, pp. 1328--1336.

\bibitem{ren2017personalized}
J.~Ren, X.~Shen, Z.~Lin, R.~Mech, and D.~J. Foran, ``Personalized image
  aesthetics,'' in \emph{International conference on computer vision}, 2017,
  pp. 638--647.

\bibitem{li2020personality}
L.~Li, H.~Zhu, S.~Zhao, G.~Ding, and W.~Lin, ``Personality-assisted multi-task
  learning for generic and personalized image aesthetics assessment,''
  \emph{IEEE Transactions on Image Processing}, vol.~29, pp. 3898--3910, 2020.

\bibitem{gatys2016image}
L.~A. Gatys, A.~S. Ecker, and M.~Bethge, ``Image style transfer using
  convolutional neural networks,'' in \emph{Proceedings of the IEEE conference
  on computer vision and pattern recognition}, 2016, pp. 2414--2423.

\bibitem{wang2017multimodal}
X.~Wang, G.~Oxholm, D.~Zhang, and Y.-F. Wang, ``Multimodal transfer: A
  hierarchical deep convolutional neural network for fast artistic style
  transfer,'' in \emph{Proceedings of the IEEE conference on computer vision
  and pattern recognition}, 2017, pp. 5239--5247.

\bibitem{liao2017visual}
J.~Liao, Y.~Yao, L.~Yuan, G.~Hua, and S.~B. Kang, ``Visual attribute transfer
  through deep image analogy,'' \emph{{ACM} Trans. Graph.}, vol.~36, no.~4, pp.
  120:1--120:15, 2017.

\bibitem{gharbi2017deep}
M.~Gharbi, J.~Chen, J.~T. Barron, S.~W. Hasinoff, and F.~Durand, ``Deep
  bilateral learning for real-time image enhancement,'' \emph{ACM Transactions
  on Graphics (TOG)}, vol.~36, no.~4, p. 118, 2017.

\bibitem{park2018distort}
J.~Park, J.-Y. Lee, D.~Yoo, and I.~So~Kweon, ``Distort-and-recover: Color
  enhancement using deep reinforcement learning,'' in \emph{Proceedings of the
  IEEE conference on computer vision and pattern recognition}, 2018, pp.
  5928--5936.

\bibitem{yan2014learning}
J.~Yan, S.~Lin, S.~Bing~Kang, and X.~Tang, ``A learning-to-rank approach for
  image color enhancement,'' in \emph{Proceedings of the IEEE conference on
  computer vision and pattern recognition}, 2014, pp. 2987--2994.

\bibitem{hu2018exposure}
Y.~Hu, H.~He, C.~Xu, B.~Wang, and S.~Lin, ``Exposure: A white-box photo
  post-processing framework,'' \emph{ACM Transactions on Graphics (TOG)},
  vol.~37, no.~2, p.~26, 2018.

\bibitem{ignatov2018wespe}
A.~Ignatov, N.~Kobyshev, R.~Timofte, K.~Vanhoey, and L.~Van~Gool, ``Wespe:
  weakly supervised photo enhancer for digital cameras,'' in \emph{Proceedings
  of the IEEE Conference on Computer Vision and Pattern Recognition Workshops},
  2018, pp. 691--700.

\bibitem{ignatov2017dslr}
A.~Ignatov, N.~Kobyshev, R.~Timofte, K.~Vanhoey, and L.~V. Gool, ``Dslr-quality
  photos on mobile devices with deep convolutional networks,'' in \emph{{IEEE}
  International Conference on Computer Vision, {ICCV} 2017, Venice, Italy,
  October 22-29, 2017}, pp. 3297--3305.

\bibitem{ying2017new}
Z.~Ying, G.~Li, Y.~Ren, R.~Wang, and W.~Wang, ``A new image contrast
  enhancement algorithm using exposure fusion framework,'' in
  \emph{International Conference on Computer Analysis of Images and
  Patterns}.\hskip 1em plus 0.5em minus 0.4em\relax Springer, 2017, pp. 36--46.

\bibitem{ke2006design}
Y.~Ke, X.~Tang, and F.~Jing, ``The design of high-level features for photo
  quality assessment,'' in \emph{Proceedings of the IEEE conference on computer
  vision and pattern recognition}, vol.~1.\hskip 1em plus 0.5em minus
  0.4em\relax IEEE, 2006, pp. 419--426.

\bibitem{bhattacharya2010framework}
S.~Bhattacharya, R.~Sukthankar, and M.~Shah, ``A framework for photo-quality
  assessment and enhancement based on visual aesthetics,'' in \emph{Proceedings
  of the 18th ACM international conference on Multimedia}.\hskip 1em plus 0.5em
  minus 0.4em\relax ACM, 2010, pp. 271--280.

\bibitem{ma2017lamp}
S.~Ma, J.~Liu, and C.~Wen~Chen, ``A-lamp: Adaptive layout-aware multi-patch
  deep convolutional neural network for photo aesthetic assessment,'' in
  \emph{Proceedings of the IEEE conference on computer vision and pattern
  recognition}, 2017, pp. 4535--4544.

\bibitem{talebi2018nima}
H.~Talebi and P.~Milanfar, ``Nima: Neural image assessment,'' \emph{IEEE
  Transactions on Image Processing}, vol.~27, no.~8, pp. 3998--4011, 2018.

\bibitem{sheng2018attention}
K.~Sheng, W.~Dong, C.~Ma, X.~Mei, F.~Huang, and B.-G. Hu, ``Attention-based
  multi-patch aggregation for image aesthetic assessment,'' in \emph{2018 ACM
  Multimedia Conference on Multimedia Conference}.\hskip 1em plus 0.5em minus
  0.4em\relax ACM, 2018, pp. 879--886.

\bibitem{DBLP}
V.~Hosu, B.~Goldl{\"{u}}cke, and D.~Saupe, ``Effective aesthetics prediction
  with multi-level spatially pooled features,'' in \emph{Proceedings of the
  IEEE conference on computer vision and pattern recognition}.\hskip 1em plus
  0.5em minus 0.4em\relax Computer Vision Foundation / {IEEE}, 2019, pp.
  9375--9383.

\bibitem{wang}
L.~Wang, X.~Wang, T.~Yamasaki, and K.~Aizawa, ``Aspect-ratio-preserving
  multi-patch image aesthetics score prediction,'' in \emph{{IEEE} Conference
  on Computer Vision and Pattern Recognition Workshops, {CVPR} Workshops 2019,
  Long Beach, CA, USA, June 16-20, 2019}, 2019, pp. 1833--1842.

\bibitem{xu2020spatial}
Y.~Xu, Y.~Wang, H.~Zhang, and Y.~Jiang, ``Spatial attentive image aesthetic
  assessment,'' in \emph{2020 IEEE International Conference on Multimedia and
  Expo (ICME)}.\hskip 1em plus 0.5em minus 0.4em\relax IEEE, 2020, pp. 1--6.

\bibitem{deng2018aesthetic}
Y.~Deng, C.~C. Loy, and X.~Tang, ``Aesthetic-driven image enhancement by
  adversarial learning,'' in \emph{2018 ACM Multimedia Conference on Multimedia
  Conference}.\hskip 1em plus 0.5em minus 0.4em\relax ACM, 2018, pp. 870--878.

\bibitem{fang2017creatism}
H.~Fang and M.~Zhang, ``Creatism: A deep-learning photographer capable of
  creating professional work,'' \emph{arXiv preprint arXiv:1707.03491}, 2017.

\bibitem{guo2018automatic}
G.~Guo, H.~Wang, C.~Shen, Y.~Yan, and H.-Y.~M. Liao, ``Automatic image cropping
  for visual aesthetic enhancement using deep neural networks and cascaded
  regression,'' \emph{IEEE Transactions on Multimedia}, vol.~20, no.~8, pp.
  2073--2085, 2018.

\bibitem{li2018a2}
D.~Li, H.~Wu, J.~Zhang, and K.~Huang, ``A2-rl: Aesthetics aware reinforcement
  learning for image cropping,'' in \emph{Proceedings of the IEEE conference on
  computer vision and pattern recognition}, 2018, pp. 8193--8201.

\bibitem{zeng2019reliable}
H.~Zeng, L.~Li, Z.~Cao, and L.~Zhang, ``Reliable and efficient image cropping:
  A grid anchor based approach,'' in \emph{Proceedings of the IEEE conference
  on computer vision and pattern recognition}, 2019, pp. 5949--5957.

\bibitem{chen2017learning}
Y.-L. Chen, J.~Klopp, M.~Sun, S.-Y. Chien, and K.-L. Ma, ``Learning to compose
  with professional photographs on the web,'' in \emph{Proceedings of the 25th
  ACM international conference on Multimedia}.\hskip 1em plus 0.5em minus
  0.4em\relax ACM, 2017, pp. 37--45.

\bibitem{wei2018good}
Z.~Wei, J.~Zhang, X.~Shen, Z.~Lin, R.~Mech, M.~Hoai, and D.~Samaras, ``Good
  view hunting: Learning photo composition from dense view pairs,'' in
  \emph{Proceedings of the IEEE conference on computer vision and pattern
  recognition}, 2018, pp. 5437--5446.

\bibitem{mnih2015human}
V.~Mnih, K.~Kavukcuoglu, D.~Silver, A.~A. Rusu, J.~Veness, M.~G. Bellemare,
  A.~Graves, M.~Riedmiller, A.~K. Fidjeland, G.~Ostrovski \emph{et~al.},
  ``Human-level control through deep reinforcement learning,'' \emph{nature},
  vol. 518, no. 7540, pp. 529--533, 2015.

\bibitem{van2016deep}
H.~Van~Hasselt, A.~Guez, and D.~Silver, ``Deep reinforcement learning with
  double q-learning,'' in \emph{Proceedings of the AAAI Conference on
  Artificial Intelligence}, vol.~30, no.~1, 2016.

\bibitem{wang2016dueling}
Z.~Wang, T.~Schaul, M.~Hessel, H.~Hasselt, M.~Lanctot, and N.~Freitas,
  ``Dueling network architectures for deep reinforcement learning,'' in
  \emph{International conference on machine learning}.\hskip 1em plus 0.5em
  minus 0.4em\relax PMLR, 2016, pp. 1995--2003.

\bibitem{sutton2018reinforcement}
R.~S. Sutton and A.~G. Barto, \emph{Reinforcement learning: An
  introduction}.\hskip 1em plus 0.5em minus 0.4em\relax MIT press, 2018.

\bibitem{mavridaki2015comprehensive}
E.~Mavridaki and V.~Mezaris, ``A comprehensive aesthetic quality assessment
  method for natural images using basic rules of photography,'' in \emph{2015
  IEEE International Conference on Image Processing (ICIP)}.\hskip 1em plus
  0.5em minus 0.4em\relax IEEE, 2015, pp. 887--891.

\bibitem{park2017personalized}
K.~Park, S.~Hong, M.~Baek, and B.~Han, ``Personalized image aesthetic quality
  assessment by joint regression and ranking,'' in \emph{2017 IEEE Winter
  Conference on Applications of Computer Vision (WACV)}.\hskip 1em plus 0.5em
  minus 0.4em\relax IEEE, 2017, pp. 1206--1214.

\bibitem{lo2012assessment}
K.-Y. Lo, K.-H. Liu, and C.-S. Chen, ``Assessment of photo aesthetics with
  efficiency,'' in \emph{Proceedings of the 21st International Conference on
  Pattern Recognition (ICPR2012)}.\hskip 1em plus 0.5em minus 0.4em\relax IEEE,
  2012, pp. 2186--2189.

\bibitem{murray2012ava}
N.~Murray, L.~Marchesotti, and F.~Perronnin, ``Ava: A large-scale database for
  aesthetic visual analysis,'' in \emph{Proceedings of the IEEE conference on
  computer vision and pattern recognition}.\hskip 1em plus 0.5em minus
  0.4em\relax IEEE, 2012, pp. 2408--2415.

\bibitem{2017Flickr}
Flickr, ``Flickr,'' \url{http://www.flickr.com}, 2017.

\bibitem{abdullah2007dynamic}
M.~Abdullah-Al-Wadud, M.~H. Kabir, M.~A.~A. Dewan, and O.~Chae, ``A dynamic
  histogram equalization for image contrast enhancement,'' \emph{IEEE
  Transactions on Consumer Electronics}, vol.~53, no.~2, pp. 593--600, 2007.

\bibitem{Thang2018Fast}
T.~Vu, C.~V. Nguyen, T.~X. Pham, T.~M. Luu, and C.~D. Yoo, ``Fast and efficient
  image quality enhancement via desubpixel convolutional neural networks,''
  2018.

\bibitem{AGCCWD}
G.~Cao, L.~Huang, H.~Tian, X.~Huang, Y.~Wang, and R.~Zhi, ``Contrast
  enhancement of brightness-distorted images by improved adaptive gamma
  correction,'' \emph{CoRR}, vol. abs/1709.04427, 2017.

\bibitem{son2019naturalness}
H.~Son, G.~Lee, S.~Cho, and S.~Lee, ``Naturalness-preserving image tone
  enhancement using generative adversarial networks,'' in \emph{Computer
  Graphics Forum}, vol.~38, no.~7.\hskip 1em plus 0.5em minus 0.4em\relax Wiley
  Online Library, 2019, pp. 277--285.

\bibitem{aestheticfatigue}
J.~S. Yu, ``On generating reasons of "aesthetic fatigue" in modernist arts--on
  the public appreciation to the works of modernist arts,'' \emph{Journal of
  Guangxi Teachers College}, 2005.

\bibitem{o2014collaborative}
P.~O'Donovan, A.~Agarwala, and A.~Hertzmann, ``Collaborative filtering of color
  aesthetics,'' in \emph{Proceedings of the Workshop on Computational
  Aesthetics}.\hskip 1em plus 0.5em minus 0.4em\relax ACM, 2014, pp. 33--40.

\bibitem{wang2016multi}
W.~Wang, M.~Zhao, L.~Wang, J.~Huang, C.~Cai, and X.~Xu, ``A multi-scene deep
  learning model for image aesthetic evaluation,'' \emph{Signal Processing:
  Image Communication}, vol.~47, pp. 511--518, 2016.

\bibitem{kao2016hierarchical}
Y.~Kao, K.~Huang, and S.~Maybank, ``Hierarchical aesthetic quality assessment
  using deep convolutional neural networks,'' \emph{Signal Processing: Image
  Communication}, vol.~47, pp. 500--510, 2016.

\bibitem{kao2017deep}
Y.~Kao, R.~He, and K.~Huang, ``Deep aesthetic quality assessment with semantic
  information,'' \emph{IEEE Transactions on Image Processing}, vol.~26, no.~3,
  pp. 1482--1495, 2017.

\bibitem{zhang2018visual}
C.~Zhang, C.~Zhu, X.~Xu, Y.~Liu, J.~Xiao, and T.~Tillo, ``Visual aesthetic
  understanding: sample-specific aesthetic classification and deep activation
  map visualization,'' \emph{Signal Processing: Image Communication}, vol.~67,
  pp. 12--21, 2018.

\bibitem{schwarz2018will}
K.~Schwarz, P.~Wieschollek, and H.~P. Lensch, ``Will people like your image?
  learning the aesthetic space,'' in \emph{2018 IEEE Winter Conference on
  Applications of Computer Vision (WACV)}.\hskip 1em plus 0.5em minus
  0.4em\relax IEEE, 2018, pp. 2048--2057.

\bibitem{kucer2018leveraging}
M.~Kucer, A.~C. Loui, and D.~W. Messinger, ``Leveraging expert feature
  knowledge for predicting image aesthetics,'' \emph{IEEE Transactions on Image
  Processing}, vol.~27, no.~10, pp. 5100--5112, 2018.

\bibitem{jin2019ilgnet}
X.~Jin, L.~Wu, X.~Li, X.~Zhang, J.~Chi, S.~Peng, S.~Ge, G.~Zhao, and S.~Li,
  ``Ilgnet: inception modules with connected local and global features for
  efficient image aesthetic quality classification using domain adaptation,''
  \emph{IET Computer Vision}, vol.~13, no.~2, pp. 206--212, 2019.

\end{thebibliography}
\end{document}